\documentclass[letterpaper]{article} % DO NOT CHANGE THIS
\usepackage[preprint]{aaai23}  % DO NOT CHANGE THIS
\usepackage{times}  % DO NOT CHANGE THIS
\usepackage{helvet}  % DO NOT CHANGE THIS
\usepackage{courier}  % DO NOT CHANGE THIS
\usepackage[hyphens]{url}  % DO NOT CHANGE THIS
\usepackage{graphicx} % DO NOT CHANGE THIS
\usepackage{natbib}  % DO NOT CHANGE THIS AND DO NOT ADD ANY OPTIONS TO IT
\usepackage{caption} % DO NOT CHANGE THIS AND DO NOT ADD ANY OPTIONS TO IT
\usepackage{algorithm}

\usepackage{subcaption}
\usepackage{enumitem}
\usepackage{xcolor}
\usepackage{newfloat}
\usepackage{listings}
\DeclareCaptionStyle{ruled}{labelfont=normalfont,labelsep=colon,strut=off} % DO NOT CHANGE THIS
\floatstyle{ruled}
\newfloat{listing}{tb}{lst}{}
\floatname{listing}{Listing}

\newcommand{\defeq}{\overset{\mathrm{def}}{=\joinrel=}}
\newcommand{\Tau}{\mathrm{T}}
\usepackage{algpseudocode}
\usepackage{amsmath}
\usepackage{amsfonts}
\usepackage{graphicx}
\usepackage{amssymb}
\newcommand{\citetApp}[1]{\citet{#1}}
\newcommand{\citeApp}[1]{\cite{#1}}

\DeclareMathOperator*{\argmin}{arg\,min}

\algrenewcommand\algorithmicrequire{\textbf{Input:}}
\algrenewcommand\algorithmicensure{\textbf{Output:}}
\usepackage{dsfont}
\newcommand{\mathbbm}[1]{\mathds{#1}}
\usepackage{booktabs}
\def\D{\mathcal{D}}
\def\L{\mathcal{L}}

\def\N{\mathcal{N}}
\def\GP{\mathcal{GP}}

\def\R{\mathbb{R}}
\def\E{\mathbb{E}}

\def\s{\boldsymbol{s}}
\def\a{\boldsymbol{a}}
\title{Co-Imitation: Learning Design and Behaviour by Imitation}
\author {
    Chang Rajani,\textsuperscript{\rm{1, 2}}
    Karol Arndt,\textsuperscript{\rm 2}
    David Blanco-Mulero,\textsuperscript{\rm 2}
    Kevin Sebastian Luck,\textsuperscript{\rm 2,3}
    Ville Kyrki \textsuperscript{\rm 2}
}
\affiliations {
    \textsuperscript{\rm 1}Department of Computer Science, University of Helsinki, Finland,\\
    \textsuperscript{\rm 2}Department of Electrical Engineering and Automation (EEA), Aalto University, Finland,\\
     \textsuperscript{\rm 3}Finnish Center for Artificial Intelligence, Finland\\
    chang.rajani@helsinki.fi, \{karol.arndt, david.blancomulero, kevin.s.luck, ville.kyrki\} @aalto.fi
}
\begin{document}

\maketitle

% TODO maybe something about unbiased estimate of measure of interest
\begin{abstract}
The co-adaptation of robots has been a long-standing research endeavour with the goal of adapting both \emph{body and behaviour} of a system for a given task, inspired by the natural evolution of animals. Co-adaptation has the potential to eliminate costly manual hardware engineering as well as improve the performance of systems.
The standard approach to co-adaptation is to use a reward function for optimizing behaviour and morphology. However, defining and constructing such reward functions is notoriously difficult and often a significant engineering effort.
This paper introduces a new viewpoint on the co-adaptation problem, which we call \emph{co-imitation}: 
finding a morphology and a policy that allow an imitator to closely match the behaviour of a demonstrator.
To this end we propose a co-imitation methodology for adapting behaviour and morphology by matching state distributions of the demonstrator. 
Specifically, we focus on the challenging scenario with mismatched state- and action-spaces between both agents.
We find that co-imitation increases behaviour similarity across a variety of tasks and settings, and demonstrate co-imitation by transferring human walking, jogging and kicking skills onto a simulated humanoid.\footnote{Find videos at \url{https://sites.google.com/view/co-imitation}}
\end{abstract}

\section{Introduction}
\label{sec:intro}
Animals undergo two primary adaptation processes: behavioural and morphological adaptation. An animal species adapts, over generations, its morphology to thrive in its environment. 
On the other hand, animals continuously adapt their behaviour during their lifetime due to environmental changes, predators or when learning a new behaviour is advantageous. 
While the processes operate on different time scales, they are closely interconnected and crucial elements leading to the development of well-performing and highly adapted organisms on earth.
While research in robot learning has largely been focused on the aspects of behavioural learning processes, a growing number of works have sought to combine behaviour learning and morphology adaptation for robotics applications via \emph{co-adaptation} \cite{luck2020data, liao2019data, schaff2019jointly, ha2019reinforcement,9701596}. 
Earlier works focused primarily on the use of evolutionary optimization techniques \cite{sims1994evolving, pollack2000evolutionary,alattas2019evolutionary}, but with the advent of deep learning, new opportunities arose for the efficient combination of deep reinforcement learning and evolutionary adaptation \cite{schaff2019jointly, luck2020data, hallawa2021evo, luck2021robot}. 
In contrast to fixed behaviour primitives or simple controllers with a handful of parameters \cite{lan2021learning, liao2019data}, deep neural networks allow a much greater range of behaviours given a morphology \cite{luck2020data}. 
\begin{figure}[t]
    \centering
    \includegraphics[width=1.0\columnwidth]{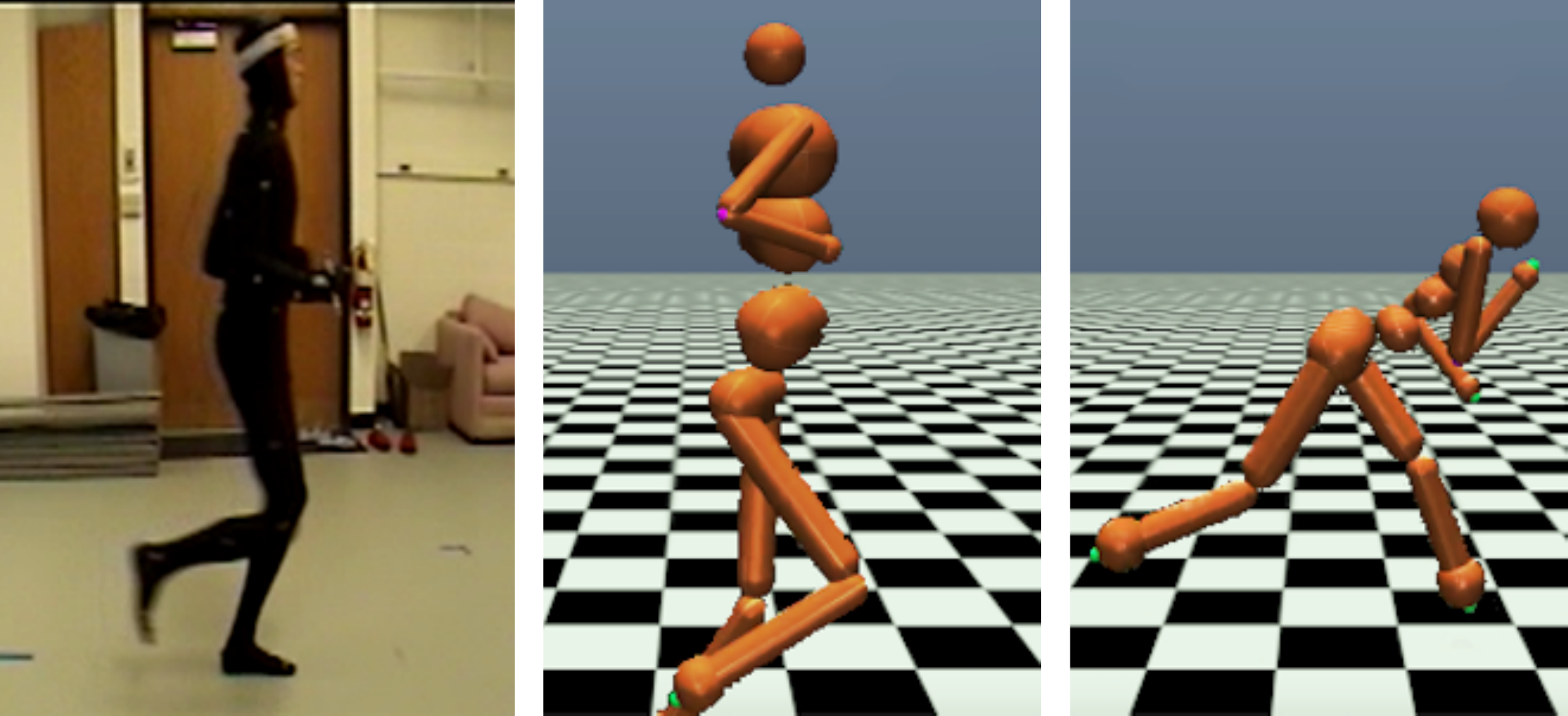}
    \caption{
    The proposed \emph{co-imitation} algorithm (\textbf{centre}) is able to faithfully match the gait of human motion capture demonstrations (\textbf{left})
    by optimizing both the morphology and behaviour of a simulated humanoid. This is opposed to a pure behavioural imitation learner (\textbf{right}) that fails to mimic the human motion accurately. 
    }
    \label{fig:hunched}
\end{figure}
Existing works in co-adaptation, however, focus on a setting where a reward function is assumed to be known, even though engineering a reward function is a notoriously difficult and error-prone task \cite{singh2019end}. 
Reward functions tend to be task-specific, and even minor changes to the learner dynamics can cause the agent to perform undesired behaviour. 
For example, in the case of robotics, changing the mass of a robot may affect the value of an action penalty.
This means that the reward needs to be re-engineered every time these properties change.
To overcome these challenges, we propose to reformulate co-adaptation by combining morphology adaptation and imitation learning into a common framework, which we name \emph{co-imitation}. 
This approach eliminates the need for engineering reward functions by leveraging imitation learning for co-adaptation, hence, allowing the matching of \emph{both the behaviour and the morphology} of a demonstrator. 
Imitation learning uses demonstration data to learn a policy that behaves like the demonstrator \cite{osa2018algorithmic,roveda2021human}.
However, in the case where the two agents' morphologies are different, we face the following challenges: (1) state spaces of demonstrating and imitating agents may differ, even having mismatched dimensionalities; (2) actions of the demonstrator may be unobservable; (3) transition functions and dynamics are inherently disparate due to mismatching morphologies. 

To address these issues we propose a co-imitation method which combines deep imitation learning through state distribution matching with morphology optimization. 
Summarized, the contributions of this paper are:
\begin{itemize}[noitemsep]
    \item Formalizing the co-imitation problem: optimizing both the behaviour and morphology given demonstration data. 
    \item The introduction of \textit{Co-Imitation Learning} (CoIL), a new co-imitation method adapting the behaviour and morphology of an agent by state distribution matching considering incompatible state spaces, without using any hand-engineered reward functions. 
    \item A comparison of morphology optimization using learned non-stationary reward functions with our proposed approach of using a state distribution matching objective.
    \item A demonstration of CoIL by learning behaviour and morphology of a simulated humanoid given real-world demonstrations recorded from human subjects in tasks ranging from walking, jogging to kicking (see Fig.~\ref{fig:hunched}).
\end{itemize}

\begin{figure*}[t]
    \centering
    \includegraphics[width=1.0\textwidth]{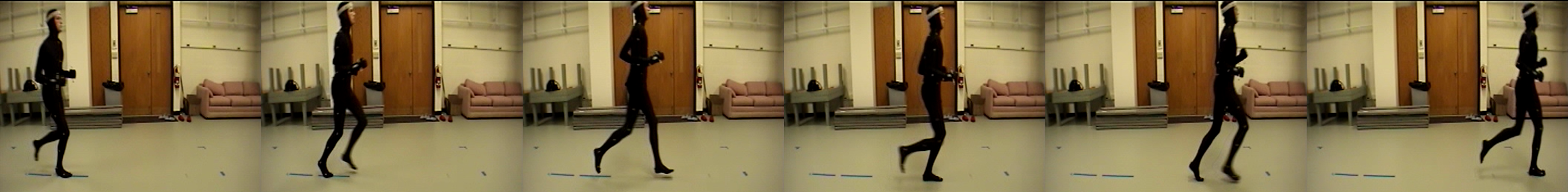}
    \includegraphics[width=1.0\textwidth]{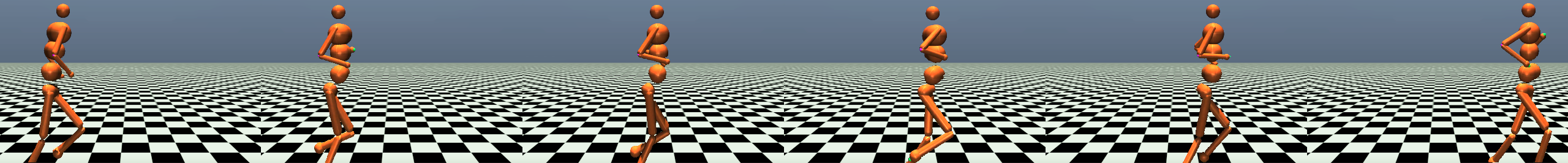}
    \includegraphics[width=1.0\textwidth]{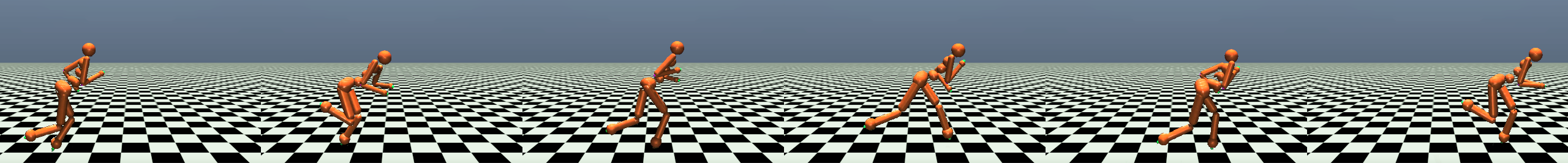}
    \caption{\textbf{Top}: A demonstrator of jogging from the CMU MoCap Dataset \cite{cmu}. \textbf{Middle}: The co-imitation Humanoid produces a more natural looking jogging motion whereas the pure imitation learner (\textbf{bottom}) learns to run with a poor gait.}
    \label{fig:co_adapt_gait}
\end{figure*}

\section{Related work }
\label{sec:related}

\paragraph{Deep Co-Adaptation of Behaviour and Design}
While co-adaptation as a field has seen interest since at least as early as the 90s \cite{park1993concurrent, sims1994evolving}, in this section we look at previous work in the field especially in the context of deep reinforcement learning. Recent work by \citet{gupta2021embodied} proposes a mixed evolutionary- and deep reinforcement learning-based approach (DERL) for co-optimizing agents' behaviour and morphology. Through mass parallelization, DERL maintains a population of 576 agents, which simultaneously optimize their behaviour using Proximal Policy Optimization (PPO) \cite{schulman2019ppo}. Based on their final task performance (i.e. episodic return), DERL optimizes the morphological structure of agents using an evolutionary tournament-style optimization process. 

\citet{schaff2019jointly} use deep reinforcement learning (RL) for the joint optimization of morphology and behaviour by learning a single policy with PPO. Again, the final episodic return of a design is used to optimize the parameters of a design distribution with gradient descent, from which the subsequent designs are sampled. Similarly, \citet{ha2019reinforcement} proposes to use REINFORCE to optimize policy parameters and design parameters of a population of agents in a joint manner. 
The co-adaptation method presented by \citet{luck2020data} improves data-efficiency compared to return-based algorithms by utilizing the critic learned by Soft Actor Critic (SAC) \cite{haarnoja2018soft} to query for the expected episodic return of unseen designs during design optimization. %, hence removing the need to maintain a population of agents. 
While the method we present is closest to the former approach, all discussed co-adaptation methods require access to a reward function, and are thus not capable of co-adapting the behaviour and design of an agent without requiring an engineer to formulate a reward function.

\paragraph{Imitation Learning with Morphology Mismatch}
Imitation learning approaches learn a policy for a given task from demonstrator data. In many cases this data can only be produced by an agent (or human) that has different dynamics from the imitator. 
We will give a brief overview on previous work where a policy is learned in presence of such transfer.
The work by \citet{desai2020imitation} discusses the imitation transfer problem between different domains and presents an action transformation method for the state-only imitation setting. \citet{hudson2022skeletal} on the other hand learn an affine transform to compensate for differences in the skeletons of the demonstrator and the imitator. These methods are based on transforming either actions or states to a comparable representation.
To perform state-only imitation learning without learning a reward function, \citet{dadashi2021primal} introduced Primal Wasserstein Imitation Learning (PWIL), where a reward function is computed based directly on the primal Wasserstein formulation.
While PWIL does not consider the case where the state space and the morphology are different between the demonstrator and the imitator, it was extended into the mismatched setting by \citet{fickinger2021cross}.
They replace the Wasserstein distance with the Gromov-Wasserstein distance, which allows the state distribution distance to be computed in mismatched state spaces.
In contrast, our method addresses the state space mismatch by transforming the state spaces to a common feature representation, allowing for more control over how the demonstrator's behaviour is imitated.
Additionally, in contrast to these works, we optimize the morphology of the imitator to allow for more faithful behaviour replication.
\citet{peng2020learning} propose an imitation learning pipeline allowing a quadrupedal robot to imitate the movement behaviour of a dog. Similarly, \citet{xu2021gan} use an adversarial approach to learn movements from human motion capture. 
Similar to us, these papers match markers between motion capture representations and robots. 
However, in the first, a highly engineered pipeline relies on a) the ability to compute the inverse kinematics of the target platform, and b) a hand-engineered reward function.
In the latter, imitation learning is used for learning behaviour, but neither method optimizes for morphology.

%===============================================================================

\section{Preliminaries}

\paragraph{Imitation Learning as distribution-matching} 
For a given expert state-action trajectory $\tau^E = (\s_0, \a_0, \s_1,$ $ \a_1, \dots, \s_n, \a_n)$, the imitation learning task is to learn a policy $\pi^I(\a | \s)$ such that the resulting behaviour best matches the demonstrated behaviour. 
This problem setting can be understood as minimizing a divergence, or alternative measures, $D(q(\tau^E), p(\tau^I \vert \pi^I))$ between the demonstrator trajectory distribution $q(\tau^E)$ and the trajectory distribution of the imitator $p(\tau^I \vert \pi^I)$ induced by its policy $\pi^I$ (see e.g. \cite{osa2018algorithmic} for further discussion). 
While there are multiple paradigms of imitation learning, a recently popular method is \emph{adversarial} imitation learning, where a discriminator is trained to distinguish between policy states (or state-action pairs) and demonstrator states \cite{ho2016generative, orsini2021matters}. The discriminator is then used for providing rewards to an RL algorithm which maximizes them via interaction.
In the remainder of the paper we will be focusing on two adversarial methods with a divergence-minimization interpretation which we will now discuss in more detail. 

\subsubsection{Generative Adversarial Imitation Learning (GAIL)}
GAIL trains a standard classifier using a logistic loss which outputs the probability that a given state comes from the demonstration trajectories \cite{ho2016generative}.
The reward function is chosen to be a function of the classifier output.
Many options are given in literature for the choice of reward, evaluated extensively by \citet{orsini2021matters}. 
Different choices of rewards correspond to different distance measures in terms of the optimization problem. 
Here, we consider the AIRL reward introduced by \citet{fu2017learning}:
\begin{align}
\label{eq:gail_reward}
    r(\s_t, \s_{t+1}) = \log(\psi(\s_t)) - \log(1 - \psi(\s_t)),
\end{align}
where $\psi$ is a classifier trained to distinguish expert data from the imitator. Maximizing the AIRL reward corresponds to minimizing the Kullback-Leibler divergence between the demonstrator and policy state-action marginals \cite{ghasemipour2020divergence}.

\subsubsection{State-Alignment Imitation Learning (SAIL)}
In contrast to GAIL, SAIL \cite{liu2019state} uses a Wasserstein-GAN-style \cite{arjovsky2017wasserstein} critic instead of the standard logistic regression-style discriminator.
Maximizing the SAIL reward corresponds to minimizing the Wasserstein distance \cite{villani2009optimal} between demonstrator and policy state-marginals (see \citet{liu2019state} for details). 
\section{A General Framework for Co-Imitation}
\label{sec:: general_coimitation}
We formalize the problem of co-imitation as follows:
Consider an expert MDP described by $(S^E,A^E, p^E, p^E_0)$, with state space $S^E$, action space $A^E$, initial state distribution $p^E_0(\mathbf{s}^E_0)$, and transition probability $p^E(\mathbf{s}^E_{t+1} \vert \mathbf{s}^E_t, \mathbf{a}^E_t)$. Furthermore, assume that the generally unknown expert policy is defined as $\pi^E(\mathbf{a}^E_t\vert\mathbf{s}^E_t)$.
In addition, an imitator MDP is defined by $(S^I,A^I, p^I, p^I_0, \pi^I, \xi)$, where the initial state distribution $p^I(\mathbf{s}^I_0\vert\xi)$ and transition probability  $p^I(\mathbf{s}^I_{t+1} \vert \mathbf{s}^I_t, \mathbf{a}^I_t, \xi)$ are parameterized by a morphology-parameter $\xi$. 
The trajectory distribution of the expert is given by
\begin{equation}
    q(\tau^E) = p_E(\mathbf{s}^E_0) \prod_{t=0}^{T-1} p_E(\mathbf{s}^E_{t+1} \vert \mathbf{s}^E_t, \mathbf{a}^E_t) \pi^E(\mathbf{a}^E_t \vert \mathbf{s}^E_t),
\end{equation}
while the imitator trajectory distribution is dependent on the imitator policy $\pi^I(\mathbf{a}\vert\mathbf{s}, \xi)$ and chosen morphology $\xi$
\begin{equation}
    p(\tau^I \vert \pi^I, \xi) = p_I(\mathbf{s}^I_0 \vert \xi) \prod_{t=0}^{T-1} p_I(\mathbf{s}^I_{t+1} \vert \mathbf{s}^I_t, \mathbf{a}^I_t, \xi) \pi^I(\mathbf{a}^I_t \vert \mathbf{s}^I_t, \xi).
\end{equation}
It follows that the objective of the co-imitation problem is to find an imitator policy $\pi^{I*}$ and the imitator morphology $\xi^*$ such that a chosen probability-distance divergence measure or function $\D(\cdot,\cdot)$ is minimized, i.e.
\begin{equation}
    \xi^*, \pi^I{}^* = \argmin_{\xi, \pi^I} \D(q(\tau^E), p(\tau^I \vert \pi^I, \xi)).
    \label{eq::co-imitation}
\end{equation}
For an overview of potential candidate distance measures and divergences see e.g. \citet{ghasemipour2020divergence}. 
For the special case that state-spaces of expert and imitator do not match, a simple extension of this framework is to assume two transformation functions $\phi(\cdot): S^E \rightarrow S^S$, and $\phi_\xi(\cdot): S^I \rightarrow S^S$ where $S^S$ is a \emph{shared feature space}.
For simplicity we overload the notation and use $\phi(\cdot)$ for both the demonstrator and imitator state-space mapping.

\section{Co-Imitation by \\State Distribution Matching}
\label{sec:coimitation-state-matching}
We consider in this paper the special case of co-imitation by state distribution matching and present two imitation learning methods adapted for the learning of behaviour and design.
The co-imitation objective from Eq.~\eqref{eq::co-imitation} is then reformulated as  
\begin{equation}
     \D(q(\tau^E), p(\tau^I \vert \pi^I, \xi)) \defeq  \D(q(\phi(\mathbf{s}^E)), p(\phi(\mathbf{s}^I) \vert \pi^I, \xi)).
\end{equation}
Similar to \citet{lee2019efficient} we define the marginal \emph{feature-space} state distribution of the imitator as
\begin{align}
\label{eq::state-dist-imitator}
    &p(\phi(\s^I)\vert \pi^I, \xi) \defeq \\
    &\E_{\substack{\s_0^I \sim p^I(\s_0^I \vert \xi) \\ \a_t^I \sim \pi^I(\a_t^I | \s_t^I, \xi) \\ \s_{t+1}^I \sim p^I(\s_{t+1}^I | \s_t^I, \a_t^I, \xi)}}
    \left[ 
        \dfrac{1}{T} \sum_{t=0}^T \mathbbm{1}(\phi(\s_t^I)=\phi(\s^I))
    \right], \nonumber
\end{align}
while the feature-space state distribution of the demonstrator is defined by
\begin{align}
\label{eq::state-dist-expert}
    &q(\phi(\s^E)) \defeq \\
    & \E_{\substack{\s_0^E \sim p^E(\s_0^E) \\ \a_t^E \sim \pi^E(\a_t^E | \s_t^E) \\ \s_{t+1}^E \sim p^E(\s_{t+1}^E | \s_t^E, \a_t^E)}}
    \left[ 
        \dfrac{1}{T} \sum_{t=0}^T \mathbbm{1}(\phi(\s_t^E)=\phi(\s^E))
    \right]. \nonumber
\end{align}
Intuitively, this formulation corresponds to matching the visitation frequency of each state in the expert samples in the shared feature space.
In principle any transformation that maps to a shared space can be used. For details of our specific choice see Section~\ref{sec:tasks}.
Importantly, this formulation allows us to frame the problem using any state marginal matching imitation learning algorithms for policy learning. See \citet{ni2021f} for a review of different algorithms.

An overview of CoIL is provided in Algorithm~\ref{alg:coil}.
We consider a set of given demonstrator trajectories $\Tau_E$, and initialize the imitator policy as well as an initial morphology $\xi_0$.
Each algorithm iteration begins with the robot training the imitator policy for the current morphology $\xi$ for $N_\xi$ iterations, as discussed in Section~\ref{sec:behaviour-adaptation}.
The set of collected imitator trajectories $\Tau^I_\xi$ and morphology are added to the dataset $\Xi$.
Then, the morphology is optimized by computing the distribution distance measure following Algorithm~\ref{alg:morpho}.
The procedure is followed until convergence, finding the morphology and policy that best imitate the demonstrator.
We follow an alternating approach between behaviour optimization and morphology optimization as proposed by prior work such as \citet{luck2020data}.
 
\begin{algorithm}
\caption{Co-Imitation Learning (CoIL)}
\label{alg:coil}
\begin{algorithmic}[1]
\footnotesize
\Require Set of demonstration trajectories $\Tau^E = \{\tau^E_0,...\}$
\State Initialize $\pi^I$, $\xi=\xi_0$,
 $\Tau^I = \varnothing$, $\Xi = \varnothing$, and RL replay $R_{\text{RL}}$
\While{not converged}
    \State Initialize agent with morphology $\xi$
    \For{$n=1, \dots, N_\xi$ episodes}
    \State With current policy $\pi^I$ sample state-action trajectory  \phantom . \phantom . \phantom . \phantom a \phantom . \phantom . \phantom . \phantom . \phantom . $(\s^I_0,\a^I_0,\dots, \s^I_t, \a^I_t,\s^I_{t+1},\dots)$   in environment
    \State Add tuples $(\s_t^I,\a_t^I,\s^I_{t+1},\xi)$ to replay $R_{\text{RL}}$
    \State Add state-trajectory $\tau^I_{n,\xi}=(\s^I_0,\s^I_1,...)$ to $\Tau^I$
    \State Compute rewards $r(\phi(\s^I_t), \phi(\s^I_{t+1}))$ using IL strategy
    \State Add rewards $r(\phi(\s^I_t), \phi(\s^I_{t+1}))$ to $R_{\text{RL}}$
    \State Update policy $\pi^I(a^I_t | s^I_t, \xi)$ using RL and $R_{\text{RL}}$
    \EndFor
    \State Add $(\xi, \Tau^I_\xi)$ to $\Xi$ with $\Tau^I_\xi=\{\tau^I_{0_\xi,\xi},...,\tau^I_{N_\xi,\xi} \}$
    \State $\xi=$ Morpho-Opt$(\Tau^E, \Xi)$ \Comment{Adapt Morphology (Alg. 2)}
  
\EndWhile
\end{algorithmic}
\end{algorithm}
\begin{algorithm}
\begin{algorithmic}[1]
\caption{Bayesian Morphology Optimization }
\label{alg:morpho}
\footnotesize
\Ensure $\xi_\text{next}$, next candidate morphology
\Procedure{Morpho-opt}{$\Tau^E$,$\Xi$}
    \State Define observations $X=\{\xi_n\}, \forall \xi_n \in \Xi$
    \State Compute $Y=\{y_n\}, \, \forall (\xi_n,\Tau^I_n) \in \Xi$ \Comment{Using Eq.~\eqref{eq:gp-targets}}
    \State Fit GP $g(\xi)$ using $X$ and $Y$
    \State $\mu_g(\tilde{\xi}), \sigma_g(\tilde{\xi}) = p(g(\tilde{\xi})\lvert X,Y)$ \Comment{Compute posterior}
    \State $\alpha(\tilde{\xi}) = \mu_g(\tilde{\xi}) - \beta \ \sigma_g(\tilde{\xi})$ \Comment{Compute LCB}
    \State $\xi_{\text{next}} = \argmin_{\tilde{\xi}} \alpha(\tilde{\xi})$  \Comment{Provide next candidate}
\EndProcedure
\end{algorithmic}
\end{algorithm}

\subsection{Behaviour Adaptation}
\label{sec:behaviour-adaptation}
Given the current morphology $\xi$ of an imitating agent, the first task is to optimize the imitator policy $\pi^I$ with
\begin{align}
    \pi^I_\text{next} =\argmin_{\pi^I} \D(q(\phi(\mathbf{s}^E)), p(\phi(\mathbf{s}^I) \vert \pi^I, \xi)).
\end{align}
The goal is to find an improved imitator policy $\pi^I_\text{next}$ which exhibits behaviour similar to the given set of demonstration trajectories $\Tau^E$. 
This policy improvement step is performed in lines 4--11 in Algorithm 1. 
We experiment with two algorithms: GAIL and SAIL, which learn discriminators as reward functions $r(\s_t, \s_{t+1})$. 
Following \cite{orsini2021matters} we use SAC, a sample-efficient off-policy model-free algorithm as the reinforcement learning backbone for both imitation learning algorithms (line 10 in Alg. 1).
To ensure that the policy transfers well to new morphologies, we train a single policy $\pi^I$ conditioned both on $s_t^I$ and on $\xi$.
Data from previous morphologies is retained in the SAC replay buffer.
Further details regarding implementation details in the setting of co-imitation is given in section \ref{appx::sec::imitation-leanring} of the Appendix.

\subsection{Morphology Adaptation}
\label{sec:morphology-adaptation}

Adapting the morphology of an agent requires a certain exploration-exploitation trade-off: new morphologies need to be considered, but changing it too radically or too often will hinder learning. In general, co-imitation is challenging because a given morphology can perform poorly due to either it being inherently poor for the task, or because the policy has not converged to a good behaviour.  
Previous approaches have focused on using either returns averaged over multiple episodes, (e.g \cite{ha2019reinforcement}) or the Q-function of a learned policy \cite{luck2020data} to evaluate the fitness of given morphology parameters. They then perform general-purpose black-box optimization along with exploration heuristics to find the next suitable candidate to evaluate. Since both approaches rely on rewards, in the imitation learning setting they correspond to maximizing the critic's \emph{approximation} of the distribution distance. 
This is because the rewards are outputs of a neural network that is continuously trained and, hence, inherently non-stationary. 
Instead, we propose to minimize in the co-imitation setting the \emph{true} quantity of interest, i.e. the distribution distance for the given trajectories. 

Given the current imitator policy $\pi^I(\a^I_t \lvert \s^I_t,\xi)$ our aim is to find a candidate morphology minimizing the objective 
\begin{align}
    \xi_\text{next} =\argmin_{\xi} \D(q(\phi(\mathbf{s}^E)), p(\phi(\mathbf{s}^I) \lvert \pi^I, \xi)),
\end{align}
using state distributions given in Eq. (\ref{eq::state-dist-imitator}) -- (\ref{eq::state-dist-expert}).

\subsubsection{Bayesian Morphology Optimization}
\label{sec:bayesian-morph-opt}
In order to find the optimal morphology parameters we perform Bayesian Optimization (BO), which is a sample-efficient optimization method that learns a probabilistic surrogate model \cite{frazier2018botutorial}.
Here, we use a Gaussian Process (GP) \cite{rasmussen_2006_gpbook} as surrogate to learn the relationship between the parameters $\xi$ and the distance $\D(q(\phi(\mathbf{s}^E)), p(\phi(\mathbf{s}^I) \lvert \pi^I, \xi))$. 
This relationship is modeled by the GP prior
\begin{equation}
    g(\xi) = \GP(\mu(\xi), k(\xi, \xi')),
\end{equation}
where $\mu(\cdot)$ defines the mean function, and $k(\cdot, \cdot)$ the kernel (or covariance) function.
We show that adapting the morphology in CoIL via this approach increases performance over the co-adaptation and imitation baselines in Section~\ref{sec:result}.

Modelling the relationship between the parameters $\xi$ and the distance $\D(\cdot,\cdot)$ is surprisingly challenging because the policy evolves over time. This means that morphologies evaluated early in training are by default worse than those evaluated later, and thus should be trusted less. The BO algorithm alleviates this problem by re-fitting the GP at each iteration using only the most recent episodes.
By learning the surrogate GP model $g(\xi)$ we can explore the space of morphologies and estimate their performance without gathering new data.
The optimization problem can be defined as
\begin{equation}
    \xi_\text{next} =\argmin_{\xi} g(\xi),
\end{equation}
where $\xi_\text{next}$ is the next proposed morphology to evaluate.
The GP model is trained using as observations the set of morphologies used in behaviour adaptation $X=\{\xi_n\}, \forall \xi_n \in \Xi$, and as targets $Y=\{y_0, \cdots, y_N \}$ the mean distribution distance for each morphology, that is
\begin{equation}
\label{eq:gp-targets}
    y_n = \dfrac{1}{N_\xi} \sum_{k=0}^{N_\xi} \D \left( q(\tau^E), p(\tau^I_{k,\xi} \lvert \pi^I,\xi) \right).
\end{equation}
The predictive posterior distribution is given by $p(g(\tilde{\xi})\lvert X,Y)=\N(\tilde{\xi} \lvert \mu_g(\tilde{\xi}), \sigma_g(\tilde{\xi}))$, where $\tilde{\xi}$ is the set of test morphologies and $\mu_g(\tilde{\xi})$ and $\sigma_g(\tilde{\xi})$ are the predicted mean and variance.
In order to trade-off between exploration and exploitation we use the Lower Confidence Bound (LCB) as acquisition function $\alpha(\tilde{\xi}) = \mu(\tilde{\xi}) - \beta \sigma(\tilde{\xi})$, where $\beta$ (here 2) is a parameter that controls the exploration.
The morphology optimization procedure is depicted in Algorithm~\ref{alg:morpho}.
The GP is optimized by minimizing the negative marginal log-likelihood (MLL).
Then, the posterior distribution is computed for the set of test morphologies $\tilde{\xi}$. The values of $\tilde{\xi}$ for each task are described in Table~\ref{table::morph_params} (Appendix).
Finally, the acquisition function is computed and used to obtain the next proposed morphology.
The Section~\ref{appx::sec::morph-optim} in the Appendix compares the proposed BO approach to Random Search (RS) \cite{bergstra2012randomsearch}, and CMA-ES \cite{hansen2001cmaes}.

\begin{figure}[t]
    \centering
    \includegraphics[width=1.0\columnwidth]{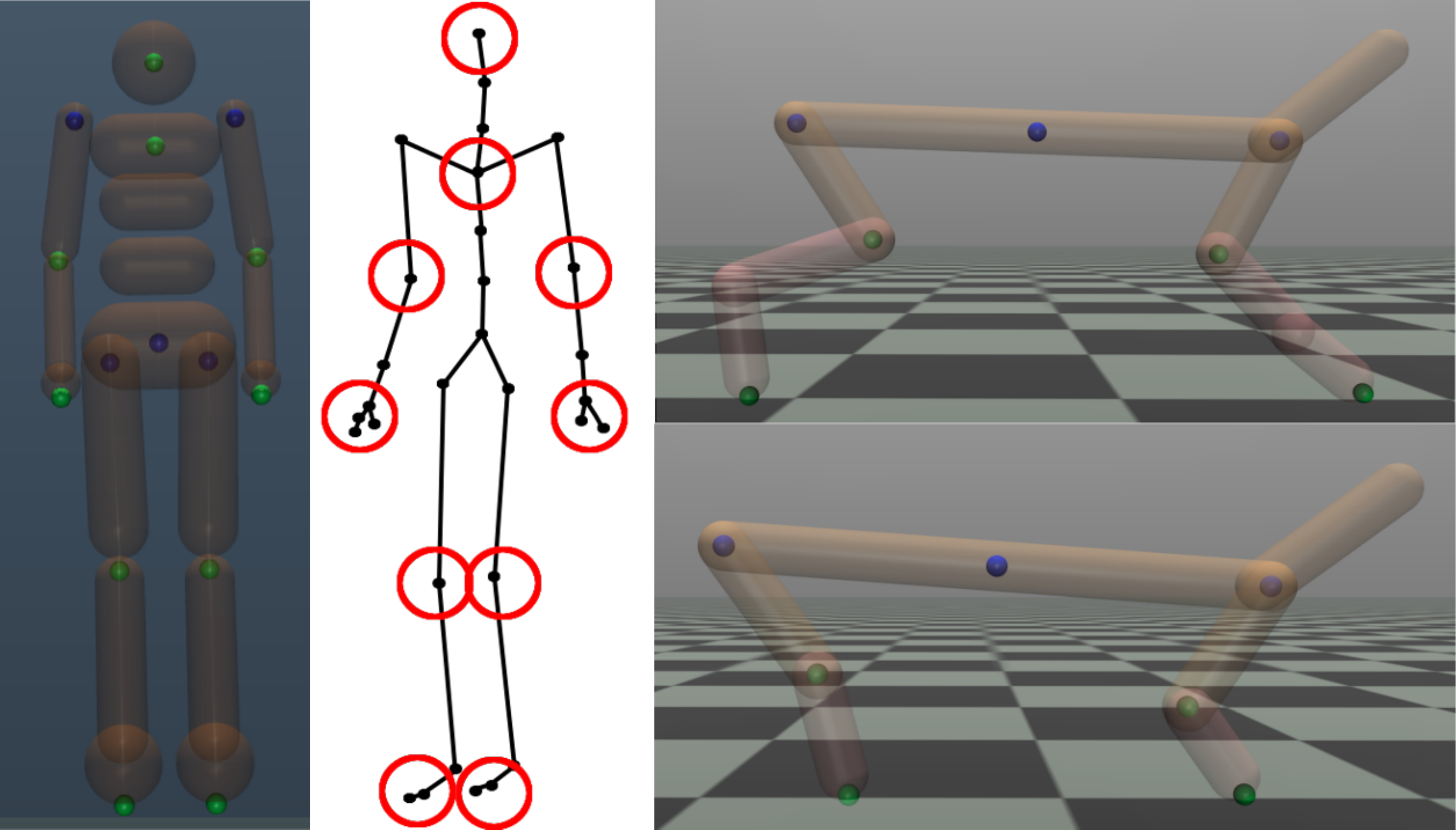}
    \caption{\textbf{Left:} Markers used for matching the MuJoCo Humanoid to motion capture data. \textbf{Right:} Markers used for the Cheetah tasks. Green markers are used as data, while blue markers serve as reference points for green markers. 
    }
    \label{fig:markers}
\end{figure}

\section{Experiments}
\label{sec:result}
Our experimental evaluation aims at answering the following research questions:
\newline
    (Q1) Does imitation learning benefit from co-adapting the imitator's morphology?
    \newline
    (Q2) How does the choice of the imitation learning algorithm used with CoIL impact the imitator's morphology?
    \newline 
    (Q3) Is morphology adaptation with CoIL able to compensate for major morphology differences, such as a missing joint or the transfer from a real to a simulated agent?
\newline
To answer these questions, we devised a set of experiments across a range of setups and imitation learning methods.

\subsection{Experimental Setup}
\label{sec:tasks}
In all our experiments, we use the MuJoCo physics engine~\cite{Todorov2012MuJoCoAP} for simulating the dynamics of agents.
As discussed in Algorithm~\ref{alg:coil}, the policies are trained using the same morphology for $N_\xi=20$ episodes. 
We optimize morphological parameters such as the lengths of arms and legs, and diameter of torso elements in humanoid (see also Table \ref{table::morph_params}, Appendix). 
The BO algorithm details as well as more detailed technical information can be found in section \ref{appx::sec::morph-bayes} (Appendix).

\subsubsection{Joint feature space $\phi(\cdot)$}

As discussed in Section \ref{sec:: general_coimitation} our method assumes that demonstrator and imitator states are in different state-spaces. 
To address this mismatch, the proposed method maps the raw state observations from the demonstrator and the imitator to a common feature space. 
The selection of the feature space can be used to influence which parts of the behaviour are to be imitated. 
In our setups, we manually selected the relevant features by placing markers along each of the limbs in both experimental setups, as shown in Figure \ref{fig:markers}). 
The feature space is then composed of velocities and positions of these points relative to the base of their corresponding limb (marked in blue in the figure).

\subsubsection{Evaluation Metric}
Evaluating the accuracy of imitation in a quantitative manner is not straightforward, because---in general---there does not exist an explicit reward function that we can compare performance on.
While most imitation learning works use task-specific rewards to evaluate imitation performance, it is not a great proxy for e.g.\ learning similar gaits. 
Recently, previous work in state-marginal matching has used forward and reverse KL divergence as a performance metric \cite{ni2021f}.
However, rather than evaluating the KL divergence, we opted for using the Wasserstein distance \cite{villani2009optimal} as the evaluation metric.
The main motivation behind this choice was that this metric corresponds to the objective optimized by SAIL and PWIL, two state-of-the-art imitation learning algorithms.
Additionally, it constitutes a more intuitive quantity for comparing 3D positions of markers than KL divergence---the Wasserstein distance between the expert and imitator feature distributions corresponds to the average distance by which markers of the imitator need to be moved in order for the two distributions to be aligned.
Therefore, for both morphology optimization and evaluation we use the exact Wasserstein distance between \emph{marker position} samples from the demonstrator $q(\phi(\s^E))$ and imitator $p(\phi(\s^I) \vert \pi^I, \xi)$ state marginal distributions. 
This also allows us to avoid an additional scaling hyperparameter when optimizing for morphologies, since velocities and positions have different scales. 
The Wasserstein distances are computed using the \emph{pot} package \cite{flamary2021pot}.
For all runs we show the mean and standard deviation of 3 seeds represented as the shaded area.

\subsection{Co-Imitation from Simulated Agents}
\begin{figure}
    \centering
    \includegraphics[width=0.99\columnwidth]{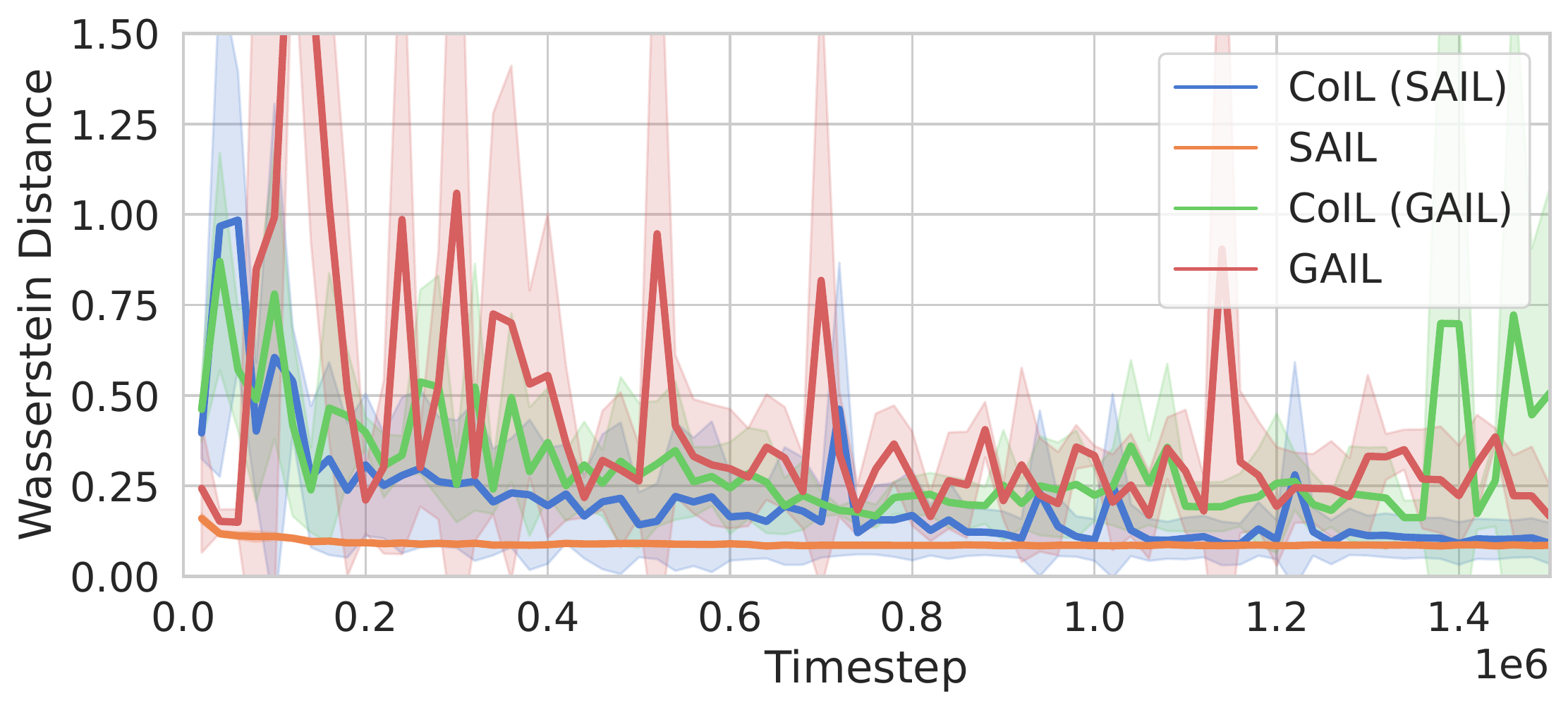}
    \caption{
    Wasserstein distance for three seeds between demonstrator and imitator trajectories on the \emph{3to2} Cheetah task on co-imitation (CoIL) and  pure imitation learning algorithms (SAIL, GAIL).
    }
    \label{fig:ablation_gail_sail_cheetah}
\end{figure}

\begin{figure}
\centering
\begin{subfigure}{0.9\columnwidth}
    \centering
    \includegraphics[width=\textwidth]{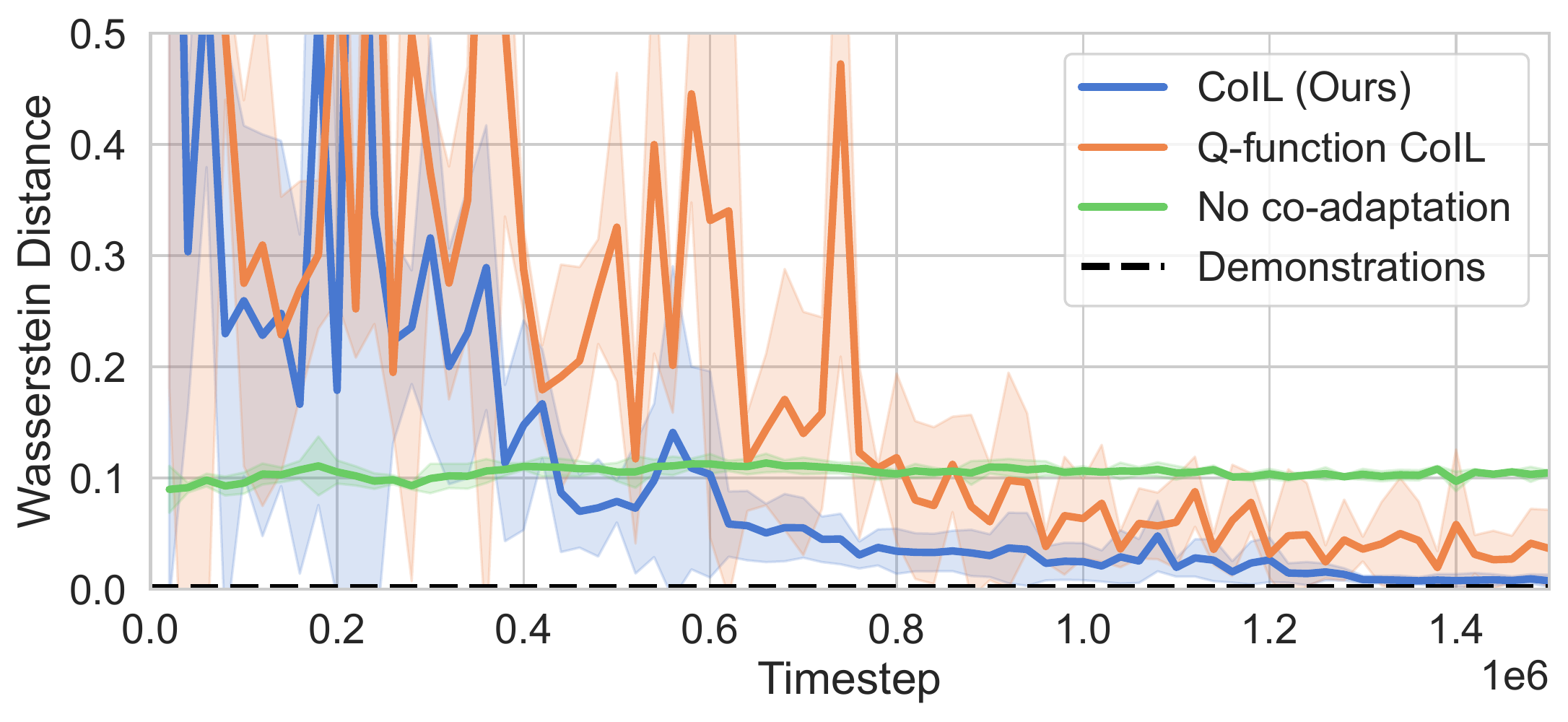}
    \caption{Imitation of a 2-joint Cheetah using a 3-joint Cheetah.}
    \label{fig:2to3_result}
\end{subfigure}

\begin{subfigure}{0.9\columnwidth}
    \centering
    \includegraphics[width=\textwidth]{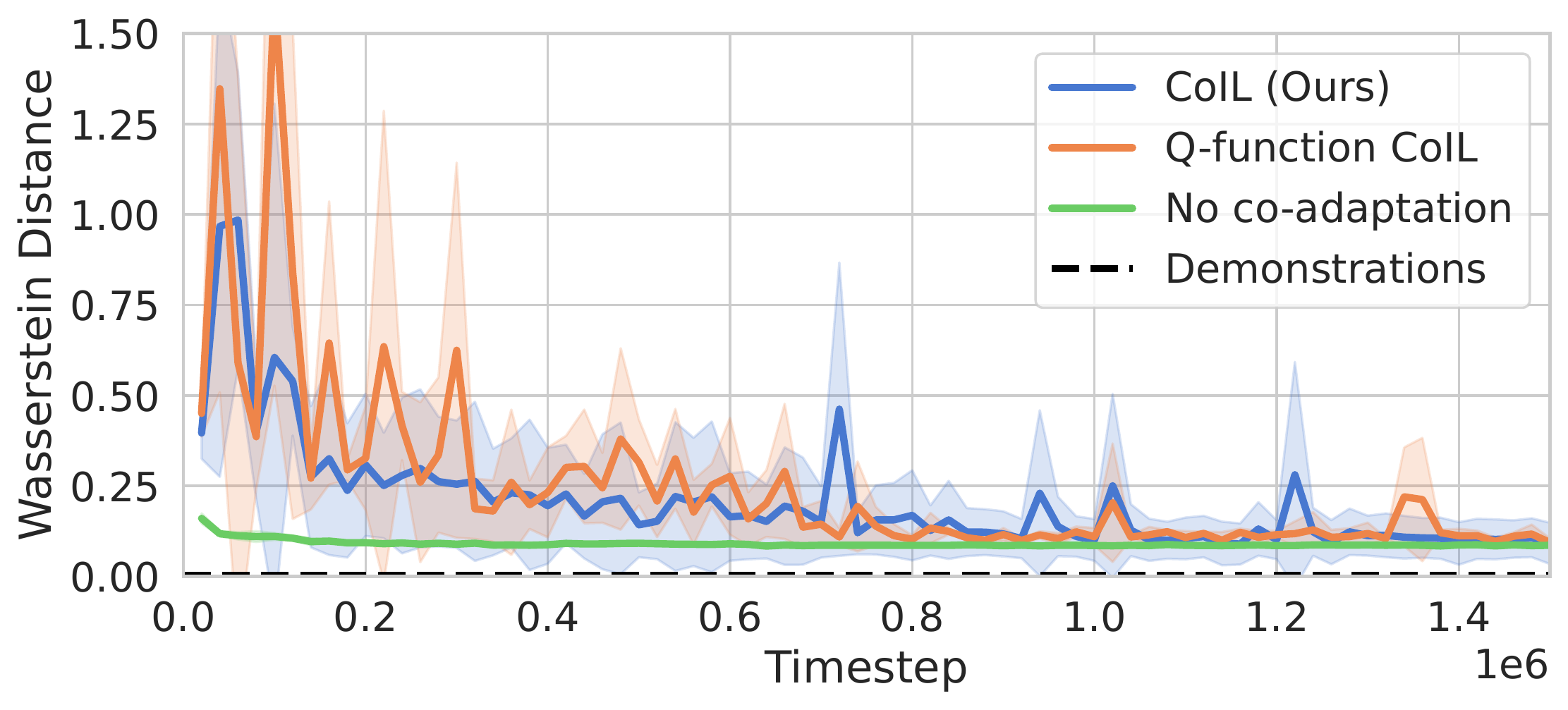}
    \caption{Imitation of a 3-joint Cheetah using a 2-joint Cheetah.}
    \label{fig:3to2_result}
\end{subfigure}
        
\caption{
Wasserstein distance for 3 seeds between demonstrator and imitator for both HalfCheetah tasks. 
While co-imitation via CoIL (blue) outperforms SAIL (green) in 2to3 (a), all methods show the same performance in 3to2 (b).
}
\label{fig:cheetah_results}
\end{figure}
\begin{figure*}[ht]
\centering
\begin{subfigure}{0.32\textwidth}
    \includegraphics[width=\textwidth]{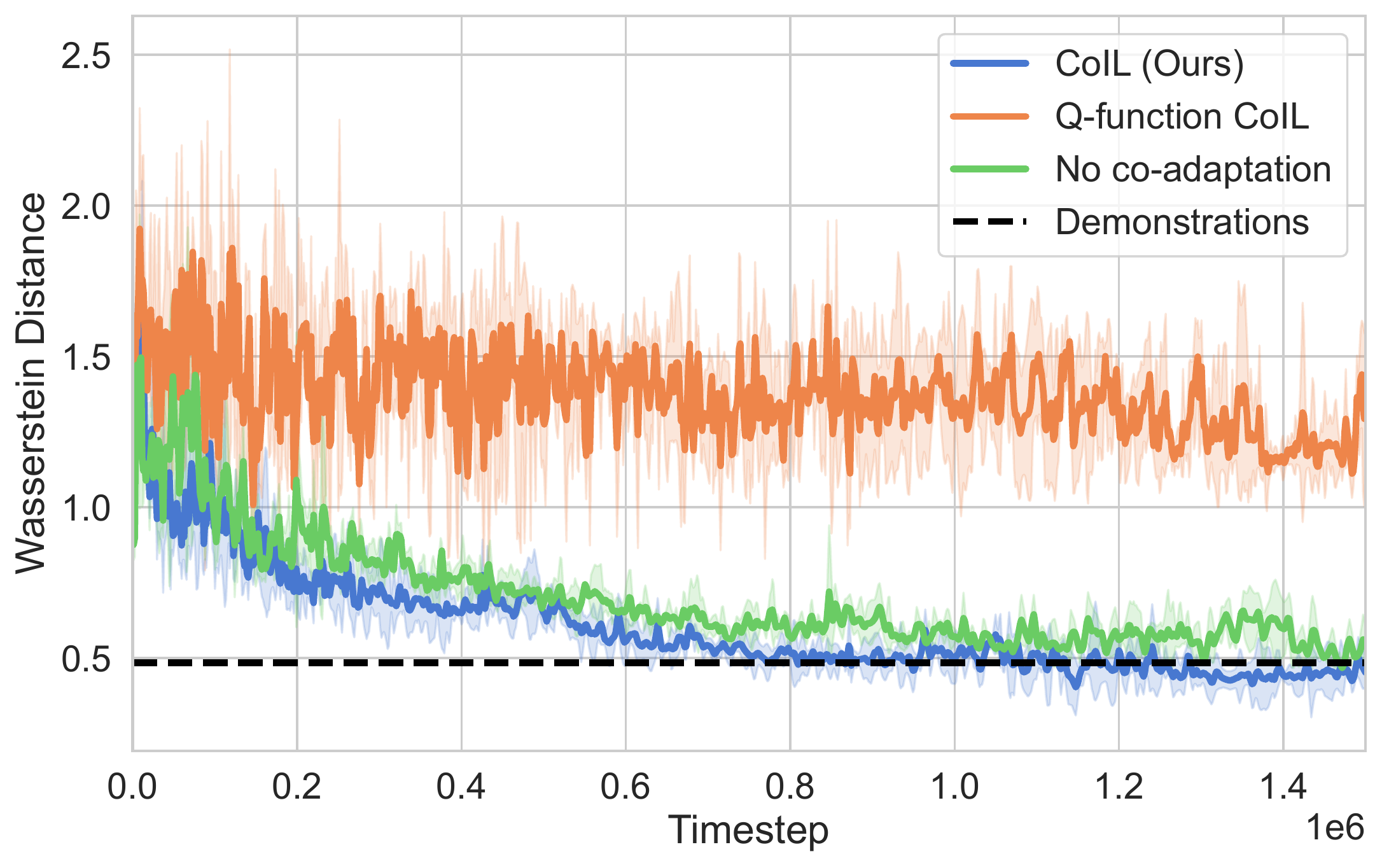}
    \caption{Soccer kick}
    \label{fig:soccer_result}
\end{subfigure}
\hfill
\begin{subfigure}{0.32\textwidth}
    \includegraphics[width=\textwidth]{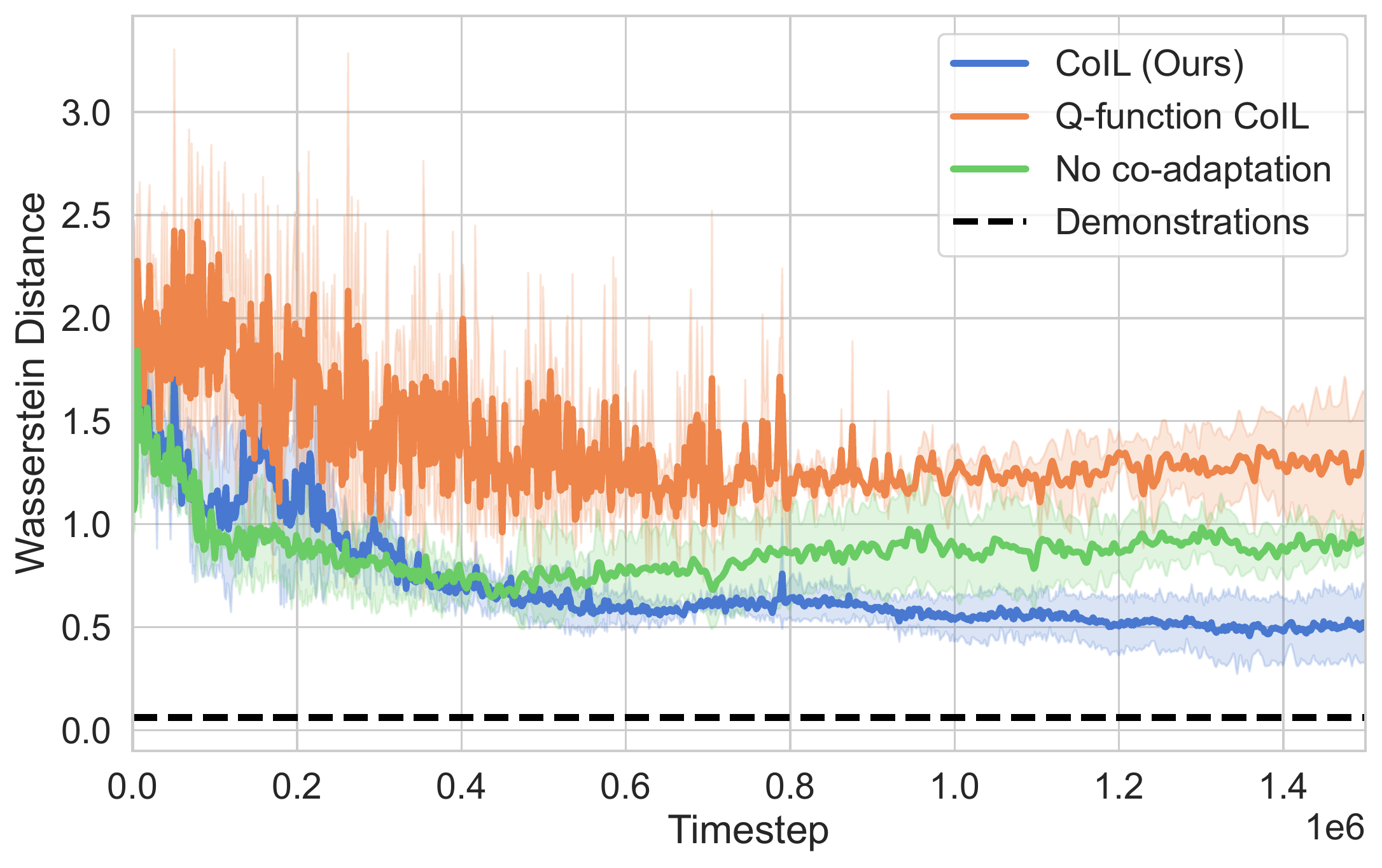}
    \caption{Jogging}
    \label{fig:jogger_result}
\end{subfigure}
\hfill
\begin{subfigure}{0.32\textwidth}
    \includegraphics[width=\textwidth]{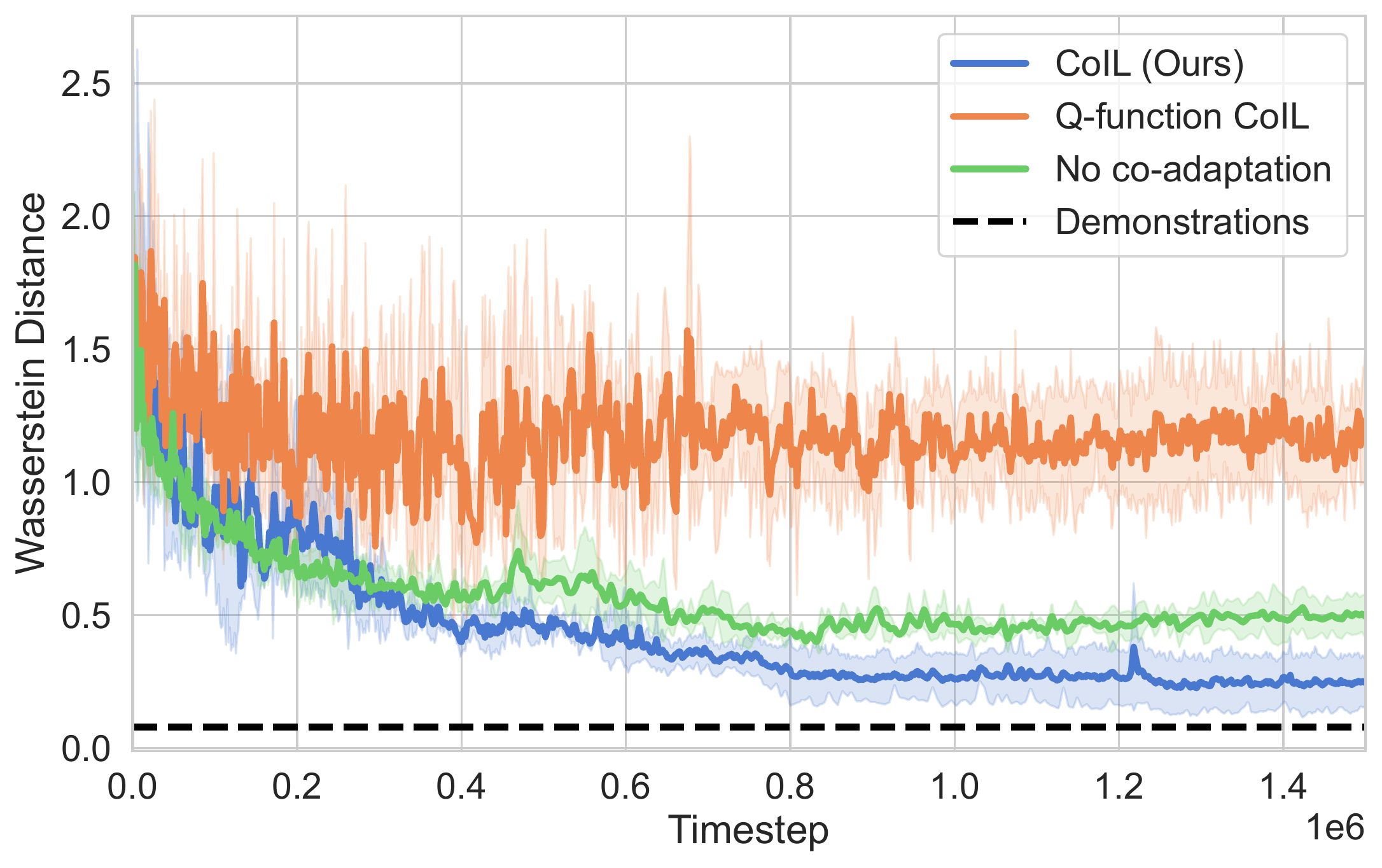}
    \caption{Walking}
    \label{fig:walker_result}
\end{subfigure}
        
\caption{The average Wasserstein distances (of 10 test episodes, 3 seeds) for the three CMU motion-capture to MuJoCo Humanoid tasks. The baseline "Demonstrations" refers to the mean distance between the individual demonstration trajectories. We can see that CoIL (blue) consistently performs better than the compared methods, even reaching the mean distance between the demonstration trajectories (black) in the soccer task.}
\label{fig:humanoid_results}
\end{figure*}
We adapt the HalfCheetah setup from OpenAI Gym~\cite{openaigym} by creating a version with two leg-segments instead of three (see Fig.~\ref{fig:markers}). We then collect the demonstration datasets by generating expert trajectories from a policy trained by SAC using the standard running reward for both variants of the environment. 
We refer to these tasks as \emph{3to2} and \emph{2to3} corresponding to imitating a 3-segment demonstrator using a 2-segment imitator and vice versa. For both experiments we used 10 episodes of 1000 timesteps as demonstration data. 
Further details can be found in the Appendix.
First, we answer RQ1 by investigating whether co-adapting the imitator's morphology is at all beneficial for their ability to replicate the demonstrator's behaviour, and---if so---how different state marginal matching imitation learning algorithms perform at this task (RQ2).
To this end, we analyze the performance of two imitation learning algorithms, GAIL and SAIL on the HalfCheetah setup, with and without co-adaptation.
We use BO as the morphology optimizer, as it consistently produced good results in preliminary experiments (see Appendix).
The performance for both imitation algorithms on the \emph{3to2} task is shown in Figure~\ref{fig:ablation_gail_sail_cheetah}.
We observe that SAIL outperforms GAIL both with and without morphology adaptation. 
Our results indicate that this task does not benefit from morphology optimization as SAIL and CoIL achieve similar performance.
However, it is encouraging to note that CoIL does not decrease performance even when the task does not benefit from co-adaptation.
Based on these results we select SAIL as the main imitation learning algorithm due to its higher performance over GAIL. 
\newline
Figure~\ref{fig:cheetah_results} shows the results in the two HalfCheetah morphology transfer scenarios.
To address RQ3, we compare CoIL to two other co-imitation approaches: using the cheetah without morphology adaptation, as well as to using the Q-function method adapted from \citet{luck2020data}.
Since this method is designed for the standard reinforcement learning setting, we adapt it to the imitation learning scenario by using SAIL to imitate the expert trajectories, and iteratively optimizing the morphology using the Q-function. See the Appendix for further details of this baseline.
In the 3to2 domain transfer scenario (Figure \ref{fig:3to2_result}), where the gait of a more complex agent is to be reproduced on a simpler setup, the results are even across the board. 
All methods are able to imitate the demonstrator well, which indicates that this task is rather easy, and that co-adaptation does not provide much of a benefit.
On the other hand, in the 2to3 scenario shown in Figure~\ref{fig:2to3_result}, after co-adaptation with CoIL, the more complex Cheetah robot is able to reproduce the gait of the simpler, two-segment robot very closely.
A closer look at the results reveals that the morphology adaptation algorithm achieves this by setting the length of the missing link in each leg to a very small, nearly zero value (see Appendix). 
Thus, at the end of training, CoIL can recover the true morphology of the demonstrator.
% \newline
While the Q-function optimization procedure \cite{luck2020data} also optimizes for the Wasserstein distance metric via the reward signal, the final performance is somewhat worse.
We hypothesize that this is due to the non-stationarity of the learned reward function and that with more interaction time, the Q-function version would reach the performance of CoIL.

\subsection{Co-Imitation from Human Behaviour}

Next, we address RQ3 by evaluating CoIL in a more challenging, high-dimensional setup, where the goal is to co-imitate demonstration data collected from a real-world human using a simplified simulated agent.
Here, we use a Humanoid robot adapted from OpenAI Gym \cite{openaigym} together with the CMU motion capture data \cite{cmu} as our demonstrations.
This setup uses a similar marker layout to HalfCheetah's, with markers placed at each joint of each limb, with additional marker in the head (see Figure \ref{fig:markers} for a visualization).
We follow the same relative position matching as in the Cheetah setup. 
We also include the absolute velocity of the torso in the feature space to allow modelling forward motion. 
The performance of the Humanoid agent on imitating three tasks from CMU motion capture dataset: \textit{walking}, \textit{jogging}, and \textit{soccer kick}, is shown in Figure~\ref{fig:humanoid_results}.
We observe that, in all three tasks, CoIL reproduces the demonstrator behaviour most faithfully. 
A comparison of the morphology and behaviour learned by CoIL vs standard imitation learning (here SAIL) in the jogging task is shown in Figure \ref{fig:co_adapt_gait}. 
In the \textit{soccer kick} task, CoIL's performance matches the distance between individual demonstrations, while for the two locomotion tasks---jogging and walking---there is still a noticeable performance gap between CoIL and the individual expert demonstrations (with $p=0.0076$, Wilcoxon signed rank test).
We also observe that, in all three setups, not performing co-adaptation at all (and using the default link length values for the OpenAI Gym Humanoid instead) outperforms co-adaptation with the Q-function objective.
We hypothesize that this counter-intuitive result might stem from the increased complexity of the task---learning a sensible Q-function in the higher-dimensional morphology- and state feature-space of Humanoid is likely to require a much larger amount of data, and thus a longer interaction time.
In contrast, optimizing the morphologies using the Wasserstein distance directly simplifies the optimization, since it does not rely on the Q-function "catching up" with changes both to policy and to the adversarial reward models used in GAIL and SAIL.

%===============================================================================
\section{Conclusion}
\label{sec:conclusion}

In this paper we presented Co-Imitation Learning (CoIL): a methodology for co-adapting both the behaviour of a robot and its morphology to best reproduce the behaviour of a demonstrator. 
This is, to the best of our knowledge, the first deep learning method to co-imitate both morphology and behaviour using only demonstration data with no pre-defined reward function. 
We discussed and presented a version of CoIL using state distribution matching for co-imitating a demonstrator in the special case of mismatching state and action spaces.  
The capability of CoIL to better co-imitate behaviour and morphology was demonstrated in a difficult task where a simulated humanoid agent has to imitate real-world motion capturing data of a human. 
Although we were able to show that CoIL outperforms non-morphology-adapting imitation learning techniques in the presented experiment using real-world data, we did not consider or further investigate the inherent mismatch between physical parameters (such as friction, contact-forces, elasticity, etc.) of simulation and real world or the use of automatic feature-extraction mechanisms. 
Limitations of CoIL are that good quality demonstrations are needed, and due to the used RL-techniques, no global optima is guaranteed. 
We think that these challenges present interesting avenues for future research and that the presented co-imitation methodology opens up a new exciting research space in the area of co-adaptation of agents. 

\clearpage
\section{Acknowledgments}
This work was supported by the Academy of Finland Flagship programme: Finnish Center for Artificial Intelligence FCAI and by Academy of Finland through grant number 328399.  We acknowledge the computational resources provided by the Aalto Science-IT project. The data used in this project was obtained from mocap.cs.cmu.edu and was created with funding from NSF EIA-0196217.

We thank the anonymous reviewers for their helpful comments and suggestions for improving the final manuscript. 
%===============================================================================

% no \bibliographystyle is required, since the corl style is automatically used.
% \bibliography{example}  % .bib
\bibliography{example}

%\newpage~\newpage~\newpage
\clearpage
\appendix
\section*{Appendix}
\section{Imitation Learning algorithms}
\label{appx::sec::imitation-leanring}
Here we describe the core RL and imitation algorithms we used in this work. For consistency and following the large ablation study conducted by \citeApp{orsini2021matters}, we used SAC \citeApp{haarnoja2018soft} as the RL algorithm for all methods. SAC is an actor-critic algorithm which optimizes a soft Q-function with the loss
\begin{align}
    \label{eq:sac_q_loss}
    \mathcal{L}(Q) = \E_{\s_t, \a_t, \s_{t+1} \sim \tau_\pi} \left[ \frac{1}{2} (Q(\s_t, \a_t) - \hat{Q})^2  \right],
\end{align}
with the target
\begin{align}
\hat{Q} = r(\s_t, \s_{t+1}) + \gamma \E_{ \a_t \sim \pi} Q(\s_{t+1}, \a_t),
\end{align}
and the policy loss
\begin{equation}
\label{eq:sac_policy_loss}
    \L(\pi) = \E_{\s_t \in \tau_\pi} \big[ \alpha_{\text{SAC}} \log \pi(\a_t | \s_t) - Q(\s_t, \a_t) \nonumber \big].
\end{equation}
including the automatic entropy tuning of $\alpha_{\text{SAC}}$ introduced in \citeApp{haarnoja2018softb}.

\paragraph{Generative Adversarial Imitation Learning}

The authors of GAIL used TRPO to maximize the adversarial reward generated by the GAIL discriminator. Adapting it to SAC is straightforward, as only the reward depends on the discriminator output.

\paragraph{State-Alignment Imitation Learning}
\label{sec:sail}

In addition to the adversarial-style rewards SAIL uses a modified policy objective which adjusts the policy towards a \emph{policy prior}
\begin{align}
    \label{eq:sail_policy_prior}
\pi_p(\a_t | \s_t) \propto \exp \left( -\left\|\frac{g_{\text{inv}}(\s, f(\s))}{\sigma}\right\|^2 \right) 
\end{align}
where $g_{\text{inv}}$ is an inverse dynamics model trained using transitions sampled from the policy, $f(\s)$ is a $\beta-$Variational Auto Encoder (VAE) trained using demonstration data, and $\sigma$ is a constant (see \citetApp{liu2019state} for details). We use 50K timesteps worth of data using random actions to pretrain the inverse dynamics and pretrain the VAE.

The authors used on-policy PPO as the base RL algorithm.
In order to use SAC, we make the following adjustments to the SAC policy objective:
\begin{align}
    \label{eq:sail_sac_loss}
    \L(\pi) = \E_{\s_t \in \tau_\pi} \big[ \alpha_{\text{SAC}} \log \pi^I(\a_t | \s_t) - Q(\s_t, \a_t) \nonumber \\ + (\pi^I(\a_t | \s_t) - \pi_p(\a_t | \s_t))^2 \big],
\end{align}
where $\alpha_{\text{SAC}}$ is the entropy scalar tuned as in \citeApp{haarnoja2018softb}. 
In this work, we include the gradient penalty term introduced in \citeApp{gulrajani2017improved} to the discriminator loss.

\section{Further Discussions of the Co-Imitation Framework}
In this section we give further intuition for matching the proposed trajectory distributions of expert and imitator by using two variations of the Kullback-Leibler (KL) divergence as example divergence to minimize. 
Our aim is to provide further intuition and discuss and shed a light at the unique properties of the selected co-imitation problems presented in this paper. 
We will start the discussion with a look at the standard imitation learning problem (i.e. behaviour cloning) in the context of co-adaptation. 
Thereafter, we discuss further the co-imitation setting selected for CoIL, namely to match the state distributions between imitator and expert. 
As a reminder, the trajectory distribution of the expert is given by
\begin{equation}
    q(\tau) = q(\mathbf{s}_0) \prod_{t=0}^{T-1} q(\mathbf{s}_{t+1} \vert \mathbf{s}_t, \mathbf{a}_t) \pi^E(\mathbf{a}_t \vert \mathbf{s}_t),
\end{equation}
while the imitator trajectory distribution is dependent on the imitator policy $\pi^I(\mathbf{a}\vert\mathbf{s}, \xi)$ and chosen morphology $\xi$
\begin{equation}
    p(\tau \vert \pi^I, \xi) = p(\mathbf{s}_0 \vert \xi) \prod_{t=0}^{T-1} p_I(\mathbf{s}_{t+1} \vert \mathbf{s}_t, \mathbf{a}_t, \xi) \pi^I(\mathbf{a}_t \vert \mathbf{s}_t, \xi).
\end{equation}
To improve readability and for the purpose of this discussion we will assume a shared state $S$ and action $A$ spaces for both imitator and expert. 
\subsection{Behavioural Cloning in the Co-Imitation Setting}
While classic imitation learning is concerned with the problem of minimizing 
\begin{align}
    \min_{\pi^I} D(q(\tau), p(\tau \vert \pi^I))
\end{align}
for a measure or divergence $D(\cdot,\cdot)$ with matching transition probabilities, we consider in this work the extension where we aim to optimize 
\begin{align}
    \min_{\pi^I, \xi} D(q(\tau), p(\tau \vert \pi^I, \xi)),
    \label{Appendix::Eq:gco:problem}
\end{align}
assuming a parameterization of the imitator transition probability of $p(\s_{t+1}\vert\s,\a,\xi)$. 
In our setting, we assume the variable $\xi$ to be morphological parameters such as lengths, sizes, weights or other shape inducing variables. 
These parameters are observable and not latent as we assume them to be needed for the instantiation of the simulation\footnote{I.e. these parameters are needed for construction of the URDF/XML files needed for simulation}, or production of the hardware components via 3D-printing, for example. 

While many potential choices for $D(\cdot,\cdot)$ exist (see e.g. \citetApp{ghasemipour2020divergence} for some options) we will provide some further intuition behind the proposed co-imitation learning problem by utilizing the KL-divergence. 
Applying the KL-divergence to the problem in Eq. \ref{Appendix::Eq:gco:problem} results in
\begin{align}
    D_{\text{KL}}\left(q(\tau) \vert\vert p(\tau \vert \pi^I, \xi)\right) = \int_{\tau} q(\tau) \ln \frac{q(\tau)}{p(\tau \vert \pi^I, \xi)} \nonumber
    \\
    = \underbrace{\E_{q(\tau)}\left[ \ln\left(\frac{q(\s_0)}{p(\s_0 \vert \xi)}\right)\right]}_{\text{match initial state distribution}}+
    \underbrace{\E_{q(\tau)}\left[\ln\left[\frac{\prod q(\s^\prime \vert \s, \a)}{\prod p(\s^\prime \vert \s, \a, \xi)} \right] \right]}_{\text{match transition distribution}} \nonumber
    \\ + \underbrace{\E_{q(\tau)}\left[ \ln\left[\frac{\prod \pi^E(\a \vert \s)}{\prod \pi^I(\a\vert\s)} \right] \right]}_{\text{match expert policy}},
    \label{Appendix::Eq:gco:kldiv}
\end{align}
where we can see that the equation can be rearranged into three problems: (1) matching the initial state-distribution, (2) matching the transition distributions, and (3) matching the policies of imitator and expert. 
For better readability the notation of $\s$ for the current state and $\s^\prime$ for the following state is used. 
The expectation is in respect to the trajectory distribution $q(\tau)$ of the expert, which in practice is replaced by a set of sampled trajectories. 
One can re-formulate Eq. \ref{Appendix::Eq:gco:kldiv} by using the distribution of state, action and next state, induced by the trajectory distribution of the expert leading to a simplified form with 
\begin{align}
    \int_{\s,\a,\s^\prime} q(\s,\a,\s^\prime) \left( \ln\left( \frac{q(\s^\prime \vert \s, \a)}{p(\s^\prime \vert \s, \a, \xi)}\right) + \ln\left( \frac{\pi^E(\a\vert\s)}{\pi^I(\a\vert\s)} \right) \right] 
    \nonumber
    \\
    \approx \E_{q(\s,\a,\s^\prime)}\left[-p(\s^\prime \vert \s, \a, \xi) - \pi^I(\a\vert\s)\right] + C,
\end{align}
where we show the resulting morphology-behaviour optimization objective plus the constant term. 
It is worth to note that thus far we have been using $\pi^I(\a\vert\s)$ for the policy distribution, i.e. a policy $\pi^I$ which does not depend on $\xi$. 
However, in practice\footnote{As we do in the proposed co-imitation learning method utilizing state-distribution matching.} one may want to utilize a policy $\pi^I(\a\vert\s,\xi)$ which is capable of predicting optimal (imitation) actions given both the current state and morphology of the imitator\footnote{Note that different actions may be optimal depending on the current morphology of the agent.}. 
Thus, leading to the alternative objective function
\begin{align}
    \E_{q(\s,\a,\s^\prime)}\left[-p(\s^\prime \vert \s, \a, \xi) - \pi^I(\a\vert\s,\xi)\right]. 
    \label{Appendix::Eq:gco:kldivobj}
\end{align}

\subsection{Co-Imitation Learning by State Distribution Matching}
While the equation above depends on the knowledge of the imitator transition distribution, an alternative to the chosen KL-divergence for trajectories as objective is to consider state-distribution matching divergences. 
Here, we want to compute a measure for 
\begin{align}
    D(q(\s\vert\pi^E), p(\s\vert\pi^I)). 
\end{align}
It is straight forward to show that knowledge of the trajectory distributions $q(\tau)$ and $p(\tau\vert \pi^I,\xi)$ allows us to derive the state distributions given the respective policy with
\begin{align}
    p(\s\vert\pi) = \int_\tau p(\s,\tau \vert \pi) = \int_\tau p(\s \vert \tau,\pi) p(\tau\vert\pi) \nonumber\\
    = \E_{p(\tau\vert\pi)}\left[ p(\s \vert \tau) \right] = \E_{p(\tau\vert\pi)}\left[ \frac{1}{T} \sum_{t=0}^T \mathbbm{1} (\s_t = \s) \right].
\end{align}
Intuitively, the state distribution $p(\s\vert\pi)$ can be computed by sampling trajectories with the policy $\pi$ and computing the expected occurrence of a state $\s$ in a trajectory $\tau$. 
This insight allows us to develop the presented co-imitation learning method using state-distribution matching without requiring access to the true transition probability $p(\s^I\vert\s,\a,\xi)$ of the imitator like in Eq. \ref{Appendix::Eq:gco:kldivobj}.  

While the main paper proposes to use a Wasserstein distance, we will present here some analysis using the KL-divergence as before. 
Using our state distributions defined above we arrive at the following objective with
\begin{align}
    &D(q(\s\vert\pi^E), p(\s\vert\pi^I)) \defeq \text{KL}(q(\s\vert\pi^E)\parallel p(\s\vert\pi^I)) \nonumber
    \\ 
     &= \footnotesize \text{KL}\left( %
     q(\s\vert\pi^E) %
     \middle\Vert \E_{p(\tau\vert \pi^I,\xi)}\left[ \frac{1}{T} \sum_{t=0}^T \mathbbm{1} (\s_t = \s) \right] \right) \\
    %  \\
     & =- \E_{q(\s\vert\pi^E)} \left[ \ln \E_{p(\tau\vert \pi^I,\xi)}\left[ \frac{1}{T} \sum_{t=0}^T \mathbbm{1} (\s_t = \s) \right] \right] + C,
\end{align}
where we move terms which do not depend on $\pi^I$ or $\xi$ into a constant $C$.
We can now rearrange this objective with
\begin{align}
\label{appendix:eq-expectations-expanded}
\footnotesize
    &-\E_{q(\s\vert\pi^E)} \left[ \ln\left( 
    \frac{1}{T} 
    \sum_{t=0}^T \E_{\s_0\sim p(\s_0\vert\xi)}[ \right.\right. \nonumber
    \\
    & \footnotesize \E_{\a_1\sim \pi^I(\a\vert\s_0,\xi)} [ 
    \E_{\s_1\sim p(\s\vert \s_0,\a_0,\xi)}[ \cdots  \nonumber
    \\
    & \footnotesize \E_{\a_{t-1}\sim \pi^I(\a\vert\s_{t-1},\xi)} [\cdots \nonumber \\
    &\E_{\s_t\sim p(\s\vert \s_{t-1},\a_{t-1},\xi)}\left[\mathbbm{1} (\s_t = \s)\right] ] \cdots ] ] \Bigg) \Bigg],
\end{align}
which computes the expected probability of being in state $\s$ at the end of a sub-trajectory unrolled for $t$ timesteps, where $\s$ is sampled from the expert state distribution. 
To simplify this objective and make it tractable for optimization we can apply Jensen's inequality and arrive at a lower bound with
\begin{align}
    - & T \cdot \text{Eq.~\eqref{appendix:eq-expectations-expanded}} \geq \sum_{t=0}^T \E_{q(\s\vert\pi^E)} \left[ \ln\left( \E_{\s_0\sim p(\s_0\vert\xi)}[ \right.\right. \nonumber
    \\
    &\E_{\a_1\sim \pi^I(\a\vert\s_0,\xi)} [ 
    \E_{\s_1\sim p(\s\vert \s_0,\a_0,\xi)}[ \cdots \nonumber
    \\
    &\E_{\a_{t-1}\sim \pi^I(\a\vert\s_{t-1},\xi)} [ \cdots \nonumber \\
    &\E_{\s_t\sim p(\s\vert \s_{t-1},\a_{t-1},\xi)}\left[\mathbbm{1} (\s_t = \s)\right] ] \cdots ] ] \Bigg) \Bigg],
    \label{Appendix::Eq:gco:state-dist}
\end{align}
where we removed the negative sign and consider this as a maximization problem in respect to the imitator policy  $\pi^I$ and imitator morphology $\xi$. 
The main insight we get from this exercise is that, unlike in the simpler behavioural cloning case discussed above in Equations \ref{Appendix::Eq:gco:kldiv} and previous, the behavioural policy $\pi^I$ and morphology $\xi$ are here inherently entangled and have to be optimized concurrently, i.e. a separation is not possible without further assumptions or simplifications. 
While in Eq. \ref{Appendix::Eq:gco:kldiv} we were able to separate the optimization problem into clearly defined components for matching transition probabilities (depending on $\xi$) and matching the expert and imitator policies, we find that this is not the case for state-distribution matching as derived\footnote{ Here for the case of the KL divergence.} in Eq. \ref{Appendix::Eq:gco:state-dist}.

\section{Experimental set-up}
\label{sec:ap:setup}

\subsection{Q-function baseline}

Here we describe how we adapt the morphology optimization procedure described in \citetApp{luck2020data} to the imitation learning setting to serve as a baseline. To do this, we use the Q-function learned by SAC on the SAIL reward as a surrogate for computing returns from entire episodes and optimizing the morphology using those returns. We use a linearly decreasing $\epsilon$-greedy exploration schedule and Particle Swarm Optimization \citeApp{eberhart1995pso} (as proposed by the original authors) to find the best morphology according to the Q-function in the case of exploitation. For exploration episodes, we sample morphology parameters uniformly from within the bounds described by table \ref{table:bounds}. The $\epsilon$ is linearly reduced over 1 million timesteps.

\subsection{Tasks}

We extract marker positions and velocities from the CMU motion capture data \citeApp{cmu} and apply preprocessing, such as resampling from 120hz to the MuJoCo speed of 66.67hz. For all tasks we include the preprocessed demonstrator data in the code supplement, as well as the preprocessing code.

\subsection{HalfCheetah tasks}

In the HalfCheetah tasks we directly optimize the lengths of each limb. The lower bound for optimization is $1\text{e-}6$ and the upper bound is twice the original value. When the imitator has three leg segments, there are a total of six parameters to optimize, while for the other case there are four. Figure \ref{fig:morpho_params_2to3} gives the learned parameters for the 2to3 task, as well as the meaning of each parameter.

\subsubsection{Humanoid tasks}

For the Humanoid tasks we optimize a scaling factor for the torso, legs and arms. The standard MuJoCo Humanoid corresponds to scaling factors $[1, 1, 1]$. Table \ref{table:bounds} gives the bounds for this optimization.
The specific motions and subject IDs are detailed in the code supplement. Due to variable amount of data available, all three Humanoid tasks have a different amount of demonstrator trajectories and all episode lengths are different. For "Soccer kick" the data includes multiple subjects.

\begin{table}
\caption{Hyper-parameters values shared throughout all experiments.}
\label{tab:hp}
    \centering
    \begin{tabular}{c|c}
        Hyperparameter & Value \\
        \toprule
        Batch size & $1024$ \\
        $\gamma$ & $0.97$ \\
        Use transitions & False \\
        disc decay & 0.00001 \\
        All networks & MLP \\
        Network layers & 3 \\
        Hidden nodes & 200 \\
        Activation & ReLU \\
        normalize obs & False \\
        Updates per step & 1 \\
        Q weight decay & $1\text{e-}5$ \\
        Optimizer & Adam \\
        Entropy tuning & True \\
        SAC $\tau$ & $0.005$ \\
        learning rate & 0.0003
    \end{tabular}
\end{table}

\begin{table}
\caption{Task-specific hyper-parameter values}
\label{tab:hp_tasks}
    \centering
    \begin{tabular}{c|c}
        Hyperparameter & Value \\
        \toprule
        Cheetah episode length & 1000 \\
        Humanoid max episode length & 300 \\
        Cheetah early termination & False \\
        Humanoid early termination & True \\ 
        Max steps & 1.5M \\
    \end{tabular}
\end{table}

\begin{table}
\caption{Q-function baseline-specific hyper-parameters values.}
\label{tab:hp_q_fn}
    \centering
    \begin{tabular}{c|c}
        Hyperparameter & Value \\
        \toprule
        Morphology Optimizer & PSO \\
        PSO particles & 250 \\
        PSO iters & 250 \\
        $\epsilon$ decay over & 1M steps 
    \end{tabular}
\end{table}

\begin{table}
\caption{SAIL-specific hyper-parameters values.}
\label{tab:hp_sail}
    \centering
    \begin{tabular}{c|c}
        Hyperparameter & Value \\
        \toprule
        VAE scaler & $1$ \\
        $\beta$ of VAE & 0.2 \\
    \end{tabular}
\end{table}

\begin{table}
\caption{GAIL-specific hyper-parameters values.}
\label{tab:hp_gail}
    \centering
    \begin{tabular}{c|c}
        Hyperparameter & Value \\
        \toprule
        Reward style & AIRL \\
        Use $\log r(\cdot)$ & True
    \end{tabular}
\end{table}

\subsection{Co-Imitation Learning Hyper-parameters}

Table~\ref{tab:hp} gives the common hyper-parameters used for all experiments, while Table~\ref{tab:hp_sail} gives SAIL-specific hyper-parameters and Table~\ref{tab:hp_gail} gives GAIL-specific hyper-parameters. Tabl\ref{tab:hp_q_fn} gives hyper-parameters for the Q-function baseline adapted from \citeApp{luck2020data}. Batch size corresponds to the size of mini-batch used for discriminator and policy updates. $\gamma$ is the discount factor. VAE scaler is the balancing term between the policy prior and the SAC policy loss (see Eq. \eqref{eq:sail_sac_loss}). Disc decay is the weight decay multiplier for the discriminator weights. Updates per step means how many policy and discriminator updates we take for each environment timestep. SAC $\tau$ is the soft target network update constant.

\subsection{Morphology  and Bayesian Optimization details}
\label{appx::sec::morph-bayes}
As described in Section~\ref{sec:bayesian-morph-opt} we use Bayesian Optimization for proposing the next candidate morphology.
As surrogate model we use a basic Gaussian Process regression model.
We use the Matern52 kernel and a constant mean function.
The kernel and mean function hyper-parameters are trained by minimizing the negative marginal log-likelihood (MLL), where the kernel function uses automatic relevance determination (ARD) \citeApp{rasmussen_2006_gpbook}. The optimization method selected is L-BFGS algorithm. The GP model, the computation of the posterior predictive distribution, as well as the Bayesian Optimization loop are implemented using GPy and GPyOpt as noted in Table~\ref{tab:soft}.

In order to compute the next morphology candidate $\xi_\text{next}$ we compute the predictive posterior distribution for the test morphologies $\tilde{\xi} \in \R^{M\times E_\xi}$, where $M$ is the number of test candidates to evaluate and $E_\xi$ is the dimension of morphology attributes.
The range of values for the test morphologies is shown in Table~\ref{table:bounds}, where the bounds column indicates the range for each of the morphology parameters in the respective environment.

The basic Gaussian Process regression model complexity is $\mathcal{O}(N^3)$, where $N$ is the number of samples used for training \citeApp{rasmussen_2006_gpbook}.
Thus, the GP suffers from poor scalability and using the entire data-set of collected imitation trajectories and morphologies $\Xi$ would highly increase the algorithm training process.
In addition, as discussed in Section~\ref{sec:morphology-adaptation} policies that were evaluated early in the training have worse performance than the recent ones, and hence, we have lower confidence on the performance of those morphologies.
For this reason, we limit the number of previous morphologies to consider to $N=200$, so that only the most recent policies and morphologies are taken into account for modeling the relationship between the distance distribution and the morphology parameters.

\begin{table}
\caption{Amount of morphology parameters $E_\xi$ and their bounds in each environment. Here "$2\text{x}$" means the optimization is bounded to twice the default size. We use (T) for torso, (L) for legs, (S) for segments and (H) for hands.}
\label{table::morph_params}
\label{table:bounds}
\centering
\hspace*{-0.5cm}
\begin{tabular}{l c c l}
\toprule
% \cline{2-5}
 Environment   & $E_\xi$ & Bounds    & Description                 \\
 \toprule
%  \cline{2-5} 
 Cheetah       & $6$       & $0 - 2\text{x}$    & Two L, three S    \\
%  \cline{2-5} 
\midrule
 2-seg Cheetah & $4$       & $0 - 2\text{x}$    & Two L, two S      \\ 
%  \cline{2-5} 
\midrule
 Humanoid      & $3$       & $0.5\text{x} - 2\text{x}$ & T, H and L scale \\
%  \cline{2-5} 
\bottomrule
\end{tabular}
\end{table}

\section{Hardware and software}

\subsection{Hardware and runtime}
We ran our experiments on a Dell PowerEdge C4140 with 2x8 core Intel Xeon Gold 6134 3.2GHz and a V100 32GB GPU.
The Humanoid experiments took around 24h to finish on this setup, while the HalfCheetah took 22h.

\subsection{Software versions used}
The experiments were ran on a cluster running CentOS 7.
The exact software packages used are reported in Table~\ref{tab:soft}
\begin{table}
\caption{Software packages used}
\label{tab:soft}
    \centering
    \begin{tabular}{c|c}
        Software & Version used \\
        \toprule
        Python & 3.8 \\
        PyTorch & 1.12 \\
        NumPy & 1.23 \\
        GPy & 1.10.0 \\
        Gym & 0.24 \\
        MuJoCo & 2.1 \\
        mujoco-py & 2.1.2 \\
        GPy & 1.10 \\
        GPyOpt & 1.2.6
    \end{tabular}
\end{table}

\section{Experimental results}

\subsection{Choice of the morphology optimizer}
\label{appx::sec::morph-optim}
In addition to the proposed Algorithm~\ref{alg:morpho} based on BO, we evaluate two other algorithms to optimise the morphology parameters: Random Search (RS) \citeApp{bergstra2012randomsearch}, and covariance matrix adaptation evolutionary strategy (CMA-ES) \citeApp{hansen2001cmaes}.
As discussed in Algorithm~\ref{alg:coil}, the policies are trained using the same morphology for $N_\xi=20$ episodes, as changing it too often would hinder learning.
For both CMA-ES and RS we follow a similar procedure where the optimization algorithms take as input the observations $X=\{\xi_n\}, \forall \xi_n \in \Xi$ and use as targets $Y=\{y_n\}, \, \forall (\xi_n,\Tau^I_n) \in \Xi$, where $y$ is computed following Eq.~\eqref{eq:gp-targets}.
As opposed to BO, we use the entire data-set $\Xi$ as data points in CMA-ES since its performance does not suffer from the number of samples. By contrast, our implementation of RS keeps the best previous morphology and randomly proposes a new morphology.

\begin{figure}
    \centering
    \includegraphics[width=0.85\columnwidth]{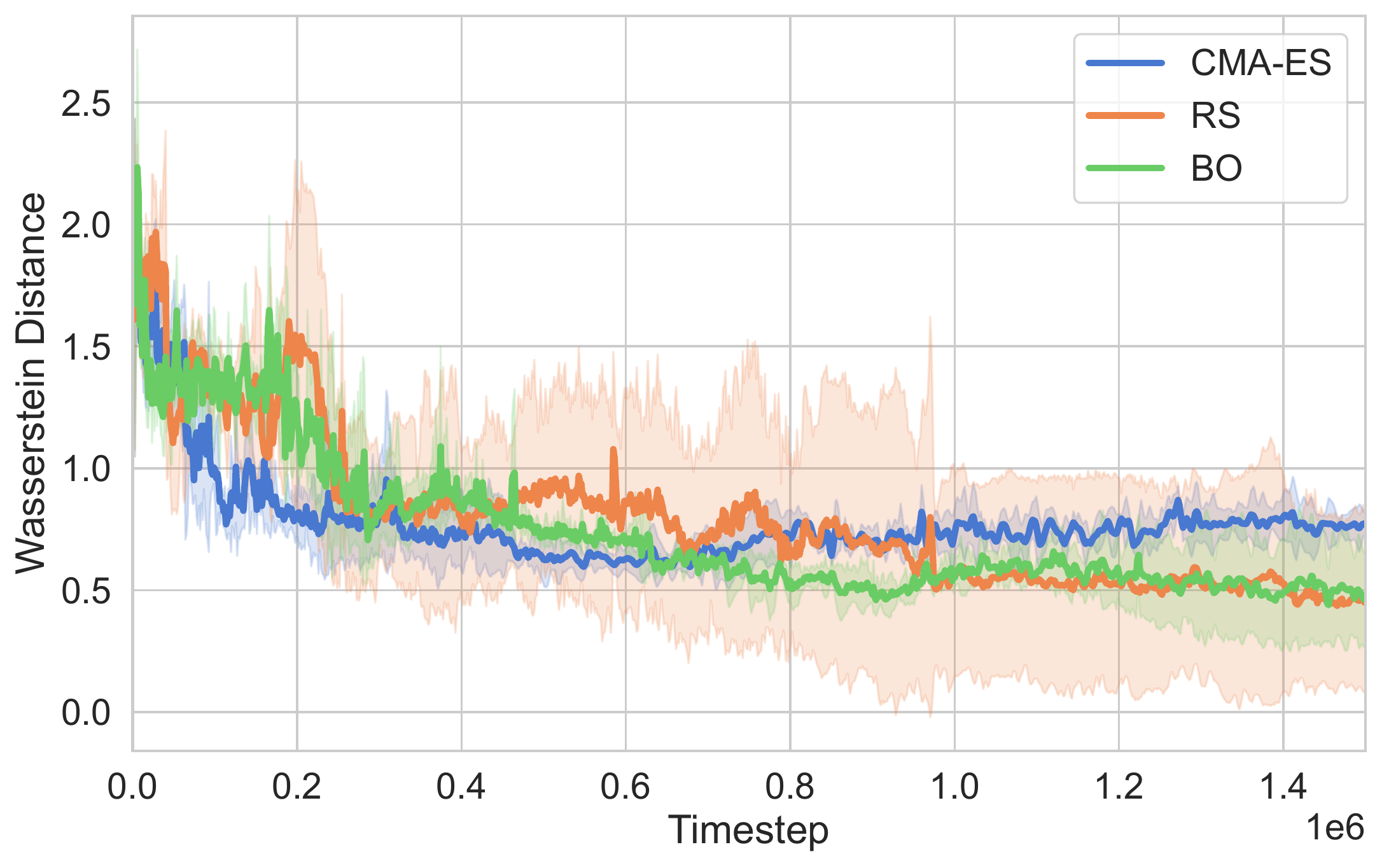}
    \caption{Wasserstein distance for the jogging experiment using CMA-ES (blue), Random Search (orange) and Bayesian Optimization (green). Both RS and BO perform comparably on average, but BO has much lower variance.}
    \label{fig:morph_optim_bo}
\end{figure}
\begin{figure*}[t]
    \centering
    \includegraphics[width=0.85\textwidth]{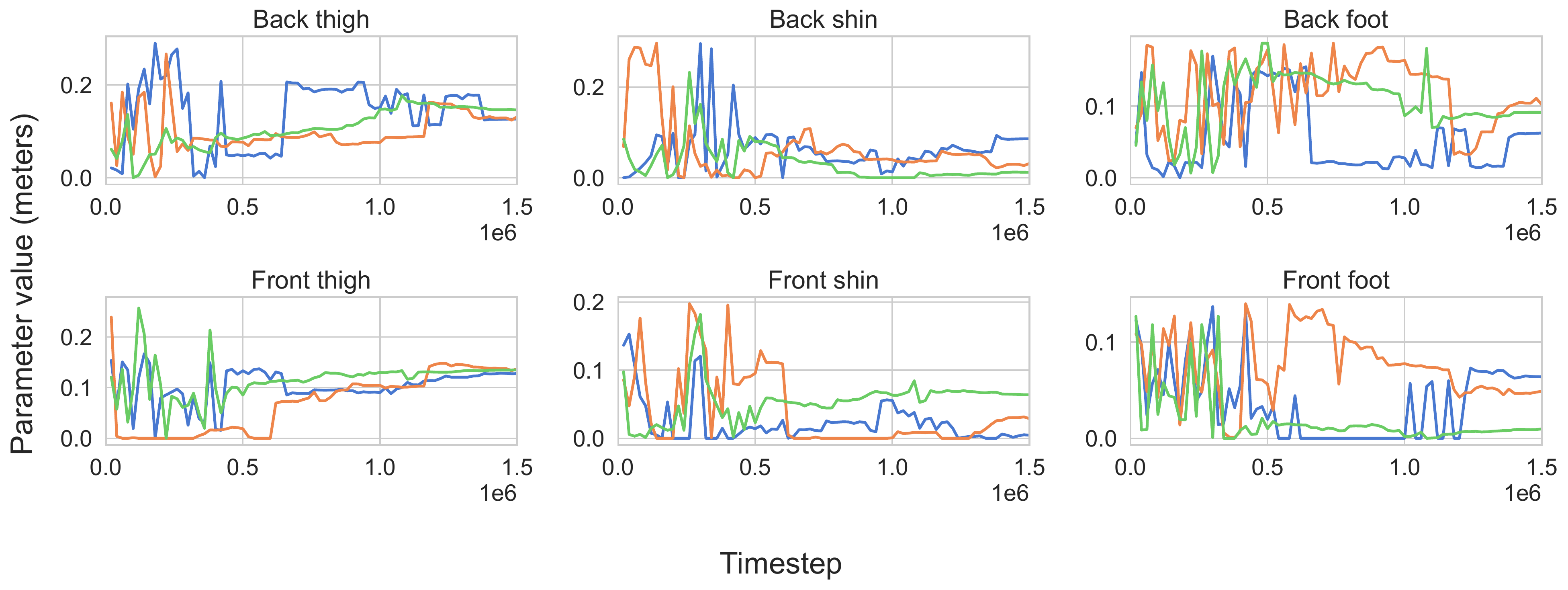}
    \caption{Morphology parameters learned by CoIL for three seeds in the 2to3 task. Note how all seeds set either the shin or the foot to close to zero which allows very close matching of the real demonstrator.}
    \label{fig:morpho_params_2to3}
\end{figure*}

The results for each optimization method in the Humanoid jogging experiment are shown in Figure~\ref{fig:morph_optim_bo}.
Since the number of different morphologies we can evaluate is relatively low, the BO approach benefits from the low data regime, presenting low mean and variance.
Similarly, the RS results present a low mean but higher variance due to the randomness inherent in the algorithm.
By contrast, we can observe that the performance of CMA-ES is lower than both BO and RS as it suffers from the low number of morphologies evaluated, getting stuck in a local optima.
These results shows that exploring the morphologies using the BO algorithm is beneficial for the task of co-imitation, as we can find an optimal solution while evaluating a low number of morphologies and keeping a relatively low variance.

\subsection{Analysis of morphology parameters}

Here we show the evolution of the morphology as a function of time for each seed of the main CoIL experiment. Figure \ref{fig:morpho_params_2to3} gives the plots for each parameter, where the parameter value is the argmax of the GP mean, i.e the best morphology so far according to the GP. 

In this task we imitate a simpler 2-leg-segment cheetah using a 3-leg-segment cheetah. It is possible for the imitator to adapt its morphology in such a way that that it matches the demonstrator exactly, by setting either both shins, both feet or one of each to zero. We can see in \ref{fig:morpho_params_2to3} that all seeds set either the shin or the foot to close to zero, meaning they are able to closely replicate the demonstrator morphology. 

\section{Limitations}
An inherent assumption of the presented approach is the availability of good demonstrations made by the expert. 
This is an inherent assumption made by the used imitation learning techniques, future work may consider the use of low-quality or incomplete sets of demonstrations and if they impact the performance of the presented co-imitation approach CoIL. 
Furthermore, the paper considers primarily continuous control problems and continuous design spaces in a reinforcement learning setting. 
Due to the use of deep reinforcement learning methods such as Soft-Actor Critic, convergence to a global optima of the presented approach cannot be guaranteed, as discussed in earlier works \citeApp{haarnoja2018soft, maei2009convergent}.
% \bibliographystyleApp{aaai23}
% \bibliographyApp{appendix}

\end{document}